\documentclass[runningheads]{llncs}
\usepackage{graphicx}
\usepackage{amsmath}
\usepackage{array}
\usepackage[caption=false,font=footnotesize]{subfig}

\usepackage{stfloats}

\usepackage[para]{threeparttable}
\usepackage{hyperref}

\usepackage{siunitx}
\begin{document}

\title{Visual Aware Hierarchy Based \\ Food Recognition}
\author{Runyu Mao\inst{1}\orcidID{0000-0002-7560-8155} \and
Jiangpeng He\inst{1}\orcidID{0000-0002-8552-9880} \and
Zeman Shao\inst{1}\orcidID{0000-0001-8088-7497} \and
Sri Kalyan Yarlagadda\inst{1}\orcidID{0000-0003-4729-6841} \and
Fengqing Zhu\thanks{Corresponding author}\inst{1}\orcidID{0000-0002-3863-3220}
}
\authorrunning{R. Mao et al.}
\institute{Purdue University, West Lafayette IN 47907, USA 
\email{\{mao111, he416, shao112, yarlagad, zhu0\}@purdue.edu}\\
}
\maketitle              % typeset the header of the contribution
\begin{abstract}
Food recognition is one of the most important components in image-based dietary assessment. However, due to the different complexity level of food images and inter-class similarity of food categories, it is challenging for an image-based food recognition system to achieve high accuracy for a variety of publicly available datasets. In this work, we propose a new two-step food recognition system that includes food localization and hierarchical food classification using Convolutional Neural Networks (CNNs) as the backbone architecture. The food localization step is based on an implementation of the Faster R-CNN method to identify food regions. In the food classification step, visually similar food categories can be clustered together automatically to generate a hierarchical structure that represents the semantic visual relations among food categories, then a multi-task CNN model is proposed to perform the classification task based on the visual aware hierarchical structure. Since the size and quality of dataset is a key component of data driven methods, we introduce a new food image dataset, VIPER-FoodNet (VFN) dataset, consists of $82$ food categories with $15k$ images based on the most commonly consumed foods in the United States. A semi-automatic crowdsourcing tool is used to provide the ground-truth information for this dataset including food object bounding boxes and food object labels. Experimental results demonstrate that our system can significantly improve both classification and recognition performance on 4 publicly available datasets and the new VFN dataset.

\end{abstract}
\section{Introduction}\label{sec:intro}

Many chronic diseases including cancer, diabetes, and heart disease, can be linked to diet. However, accurate assessment of dietary intake is an open and challenging problem. Conventional assessment methods including food records, 24-hour dietary recall, and food frequency questionnaires (FFQ) are prone to biased measurement and are burdensome to use~\cite{livingstone2004,rockett2003evaluation,larsson2002validity,poslusna2009misreporting,kirkpatrick2014performance} . In the last decade, there has been a growing popularity of using mobile and wearable computing devices to monitor diet-related behaviors and activities~\cite{zhu2010A,sun2014ebutton,Willsense}. At the same time, advances in computer vision and machine learning has enabled the development of image-based dietary assessment systems~\cite{zhu-2015,murphy_2015,fang-nu2019,he2020multitask} that can analyze food images captured by mobile and wearable devices to provide an estimate of dietary intake. Accurate estimation of dietary intake relies on the system's ability to distinguish foods from the image background (i.e., localization), to identify (or label) food items (i.e., classification), to estimate food portion size, and to understand the context of the eating event. Although there are promising advancements, many challenges still remain in automating the assessment of dietary intake from images. 

Recently, a series of breakthroughs in computer vision have been led by deep learning~\cite{dlnature} techniques such as Convolutional Neural Networks (CNNs)~\cite{krizhevsky2012imagenet} using data-driven approaches. The success of deep learning methods depends largely on the quantity and quality of data. The increased availability of online food images shared on social media and review forum has made the collection of large food image dataset possible. However, high quality ground-truth labels are equally, if not more important, for improving the performance of image-based dietary assessment systems.
In this paper, we selected the most frequently consumed foods in the United States~\cite{wweia} as the food categories to build a new food image dataset, Viper FoodNet (VFN) Dataset. A semi-automatic image collection system and a web-based crowdsourcing tool \cite{Shao2019} are implemented to collect and annotate those food images for the VFN Dataset.

Food classification~\cite{hand_crafted2,yang2010,Kagaya,structured,wu2016learning,wide} is the task of labeling food items in an image, which assumes the input image contains only a single food item. Thus, there is no need to output the pixel location of the foods in an image. Others have worked on food detection~\cite{aizawa_2013,Kagaya,highly}, which determines whether an image contains food or not. However, it is common for food images to contain multiple foods. In this paper, we define food recognition as the task of simultaneously localize and label foods present in an image. This task also provides value to subsequent tasks such as estimating food volume and portion size.
Recently, \cite{Bolaos2016SimultaneousFL} proposed a method to generate bounding boxes for food localization. However, the process is quite complex and many hand-crafted parameters are required, it is difficult to generalize to larger foods datasets or real-life food images.

In this work, we propose a two-step food recognition system that can localize and label multiple foods in an image and generalize to other food image datasets. Our system consists of food localization and food classification to handle multi-food images which are prevalent in daily life. Since many single food images also contain non-food objects such as human hands, menus, and tables, a well-designed food localization process can also remove background clutters to improve classification performance. Given the success of Faster R-CNN~\cite{ren2015faster} for face detection~\cite{jiang2017face} and hand detection~\cite{deng2017joint}, we implemented a version of it for the food localization task.

Labeling food categories for each region selected by food localization is a typical classification task. Many existing methods~\cite{huang2017densely,He_2016_CVPR} have shown good performance for image classification of general objects.
However, different food categories contain many hidden relations. For example, pancake and waffle are closely related compared to pancake and pork chop based on semantic meaning. In~\cite{wu2016learning}, instead of treating food classification as a flat fine-grained classification problem, a semantic-based hierarchical structure for food categories is proposed to improve the performance of classification. However, it is time-consuming and impossible to manually build such a hierarchical structure given that there are thousands of food categories in the world. It is also impractical to build a specific hierarchical structure for each available food image dataset since the same food may have different names or the same name may represent different foods in different regions of the world. In this paper, we proposed a method to automatically cluster visually similar foods and build a hierarchical structure for the food categories. We classify food regions by leveraging the hierarchical structure in a multi-task manner. 

In summary, the contribution of this paper is twofold. We developed an image-based food recognition system consisting of food localization and hierarchical food classification, which can be used as building blocks for image-based automatic dietary assessment applications. We designed and constructed a new food image dataset, VIPER-FoodNet (VFN) dataset, which contains 15k real-life multi-food images of most frequently consumed foods in the United State. This dataset along with other publicly available food image datasets are used for evaluating the performance of our food recognition system. 

The rest of the paper is organized as follows. In Section~\ref{Related_work}, we discuss and summarize related works to different components of our system. We discuss the design and construction of the VFN dataset in Section~\ref{Dataset}. Details of the food recognition system are presented in Section~\ref{recognition}. In Section~\ref{Experiment}, we discuss experimental results on different datasets and compare our method to other food localization and classification methods. Discussion and Conclusion is provided in Section~\ref{Discussion} and~\ref{Conclusion}.

\section{Related Work}\label{Related_work}
\subsection{Image-based Dietary Assessment}
With the development of mobile and wearable technologies, a wide range of approaches using food images captured at the eating scenes have been proposed to assess dietary intake~\cite{boushey_2017new}. Many image-based approaches have been integrated into sensor systems and mobile devices with the goal of automating dietary assessment by reducing human input, e.g., mFR~\cite{zhu2010A}, eButton~\cite{sun2014ebutton}, FRapp~\cite{Casperson2015AMP}, NuDAM~\cite{rollo2015evaluation}, and \cite{360}. The broad range of approaches can be classified into two main categories: active capture and passive capture.

\textbf{Active capture} requires the participant or a data collector to operate the device and capture images of food being consumed. A mobile device is typically used for image capturing. In addition to images, some methods include other forms of inputs. For example, FRapp~\cite{Casperson2015AMP} captures text and voice as additional information. NuDAM~\cite{rollo2015evaluation} includes voice recording from the participant about the food consumed, leftovers, information about the meal occasion, and a follow-up call in the next day for making the adjustment. On the contrary, the mFR~\cite{zhu2010A,ahmad2016mobile} provides an end-to-end solution based on capturing a pair of before and after eating scene images from a mobile device camera as the sole input to estimate food consumption using image analysis techniques~\cite{zhu-2015,Wang2017,fang-nu2019,wang2018context}. 

\textbf{Passive capture} takes pictures or videos automatically in real time or semi-automatically at intervals. Compared to active capture, it may be more challenging to process collected data due to uncertainty of how food consumption is captured, and possible concerns about privacy. The eButton~\cite{sun2014ebutton} is a chest-worn device that takes images at a preset rate during eating occasions. Food-specific models are used to estimate food volume~\cite{Jia20123DLO}. Another approach~\cite{360} estimates food intake based on video of the eating scenario captured by a 360 camera. Mask R-CNN~\cite{mask_rcnn} is used to detect people and 13 food categories.

\subsection{Food Detection and Food Localization}
Food detection, which refers to the presence of foods in an image, is a binary classification problem~\cite{Kagaya,Ragusa2016,Singla2016,Farinella2015,highly,foodlogA}. On the other hand, Food localization aims to detect where the food is located in an image, commonly indicated by bounding boxes or pixel-level segmentation masks. It is also known as food region detection~\cite{Miyano2012FoodRD}. To avoid terminology confusion, we use food localization throughout this paper with more details about our method described in Section~\ref{Localization}. Although we can extract spatial information about the foods in the image, localization will not return food categories associated with the spatial information. In this section, we will illustrate previous works of both food detection and food localization.

\textbf{Food detection,} traditionally, is solved based on handcrafted features, e.g., scale-invariant feature transform (SIFT)~\cite{Lowe2004} and speeded up robust features (SURF)~\cite{bay2008} for feature representations. 
Support vector machine (SVM)~\cite{cortes1995support} is then used to do binary classification. 
A food logging system proposed in~\cite{foodlogA} performs automated food/non-food classification based on global features~\cite{FeiFei2005aa}, e.g. circles detected by Hough transforms and the average value in color space. In 2015, a one-class SVM~\cite{chang2011} for food detection was proposed in \cite{Farinella2015}. 
Without non-food images in the training dataset, this method still achieved promising performance. Given the success of Convolutional Neural Networks (CNNs) for image classification, authors in \cite{Kagaya,Farinella2015,highly,Singla2016} fine-tuned existing CNN models for food detection by treating it as a binary classification task. 

\textbf{Food localization} requires more information since its goal is to locate regions in a food image that corresponds to food. Local handcraft features~\cite{Swain1991ColorI} are adopted since it can provide more spatial information than global feature representation \cite{FeiFei2005aa}. In \cite{Miyano2012FoodRD}, the Bag-of-Features representation~\cite{joutou2009} and local color features~\cite{Swain1991ColorI} are used to generate feature representations of the input image. The 2-class SVM~\cite{cortes1995support} and Histogram Intersection~\cite{Swain1991ColorI} is then trained and applied for food localization. 
Class Activation Map (CAM)~\cite{zhou2015cnnlocalization} proposed in 2016 showed that CNNs trained for classification can also coarsely highlight objects' positions in the image. A food activation map (FAM) \cite{Bolaos2016SimultaneousFL} is proposed in 2017, which is a kind of CAM that is sensitive to food images based on modified GoogLeNet~\cite{szegedy2015going} for food localization.

\subsection{Food Image Recognition}
Food recognition plays a key role in image-based dietary assessment. The goal is to automatically detect pixels in an image corresponding to foods and label the type of foods. Food image recognition is similar to the task of object detection. However, it is much more challenging since it requires fine-grain recognition of different foods. Moreover, many foods have similar visual appearance and foods are generally non-rigid. There are two main categories of food recognition: single-food recognition and multiple-food recognition.

\textbf{Single food recognition} assumes that only one food is present in the image. Therefore, the problem can be viewed as food image classification. Unlike the general image classification problem, food classification is much more challenging due to intra-class variation and inter-class confusion~\cite{wide}. Also, depending on personal preference, recipe used, and the availability of ingredients, the same food may have very different visual appearance. On the other hand, using the same cooking method may cause different foods to have a similar appearance, e.g., fried chicken breast and fried pork.

Earlier work~\cite{hand_crafted2} focuses on fusing different image features including SIFT, Gabor, and color histograms to classify food categories. Due to the success of using CNNs for feature representation, several groups have proposed different CNN backbone models that are pre-trained on ImageNet dataset~\cite{ilsvrc_15}, and fine-tuned these models for food image classification. For example, AlexNet~\cite{krizhevsky2012imagenet} is used in ~\cite{Kagaya}, the Network in Network model~\cite{lin2013network} is fine-tuned in ~\cite{Tanno2016}, and the Wide residual networks~\cite{zagoruyko2016wide} is modified in~\cite{wide}. Although the performance of food classification has been improved since, these methods use the same underlying concept, that is relying on one CNN model to extract image feature and perform classification. 

In 2016, Wu {\it et al} proposed a new concept that uses semantic hierarchy to learn the relationship between food categories \cite{wu2016learning} . The advantage of this method is the classifier can make a better mistake, which has closer semantic meaning to the actual food category. However, the semantic hierarchical structure is manually generated and is designed for a specific dataset. Therefore, this hierarchical structure cannot be generalized to different food image datasets. In addition, the same food may have different names or the same name may represent different foods in different cultures. Thus, building a good and adaptive semantic hierarchical structure is quite challenging. 

\textbf{Multiple food recognition} analyzes food images containing multiple foods which are more close to real-life scenario. Matsuda {\it et al} proposed a two-step solution which consists of regions proposal and region classification based on hand-crafted features ~\cite{Matsuda:2012ab}. 
Following this concept, a CNN-based two-step recognition system is developed in \cite{CNNfood} which is assisted by Selective Search~\cite{uijlings2013selective}, bounding box clustering~\cite{CNNfood}, and GrabCut~\cite{rother2004grabcut}.
The Selective Search and bounding box clustering are used to propose region candidates.
Saliency maps corresponding to candidate proposals are estimated from a pre-trained CNN model by back-propagating loss to input images.
Segments, generated by GrabCut based on saliency maps, were unified and integrated.
A bounding box is adjusted to one segment as a final region proposal and the region inside is provided to another CNN model to generate a class label.
The performance of this method is limited by the threshold used for the saliency map, and the quality of segmentation generated by GrabCut.

CNN features are used to generate a Food Activation Map (FAM) in \cite{Bolaos2016SimultaneousFL}, which places more weights to the food regions. The candidate regions are proposed based on the FAM and are fed to the other CNNs for classification. The size of FAM is $14 \times 14$. Although it can provide some spatial information to localize foods, much useful information may be lost using such a low resolution activation map. In addition, the food region generation depends on three parameters manually determined by cross-validation on UEC-256~\cite{kawano2014automatic} and EgocentricFood~\cite{Bolaos2016SimultaneousFL}. Thus, it is difficult to generalize the performance of the method on other datasets. 

Unlike the two-step method mentioned above, YOLO v2~\cite{Redmon_2017_CVPR} is used in \cite{Aguilar2017GrabPA} to recognize multiple foods in an image. To further improve the result, segmentation masks generated by the Fully Convolutional Neural Network (FCN) and the Non-maximum Suppression technique are used to remove wrong predictions. However, the test dataset, UNIMIB2016~\cite{cioccaJBHI}, used in this work has fixed settings where each image includes a tray with several foods placed on plates. 
Faster R-CNN~\cite{ren2015faster} was adopted in~\cite{2017estimating} for multiple food recognition. It is also a one-step method where the Faster R-CNN generates both bounding boxes and food labels simultaneously. The UEC-100 dataset~\cite{Matsuda:2012ab} is used for evaluation. However, the Faster R-CNN training process is not supervised by multiple food images since only single food images are used for training. The multiple food images, which is a small portion (~9.2\%) of UEC-100, are used for inference.

\subsection{Food Image Datasets}
The success of deep learning methods depends largely on the quantity and quality of data.
Therefore, a large quantity and good quality of training data would ,in general, improve the accuracy of training-based methods.
Several publicly available food image datasets, as shown in Table~\ref{tab:dataset_compare}, have been widely used to assess dietary intake based on deep learning approach. 

Table~\ref{tab:dataset_compare} summarizes the characteristics of different datasets for both single food recognition and multiple food recognition.
We report the size of each dataset (i.e. number of food images and food categories). To better understand the advantages and limitations of each dataset, we also report the food types included in the dataset and collection method (e.g., \textbf{Free-living} for unconstrained image capturing environment, or \textbf{Controlled} settings for fixed lighting conditions, dinnerware such as plates, glasses, and silverwares).
Many datasets do not specify the food type selection process and use constrained environments for image acquisition. 
The annotation can be categorized into three types: image-level label only, bounding boxes, and polygonal areas. Datasets contain image-level label only are designed for single food classification. For multiple food recognition datasets, we also report the percentage of multi-food images.

Although these datasets contain a large number of food images, they do not always meet the needs of different applications. 
The food categories in PFID~\cite{chen2009} and Menu-Match~\cite{beijbom2015menu} are limited to popular fast foods and foods from specific restaurant menus.
Food-101~\cite{bossard14} and UPMC-101~\cite{wang2015recipe} contain many noisy images which have incorrect food labels.
Recipe1M~\cite{recipe1m}, Recipe1M+~\cite{recipe1m+} , and Vireo-172~\cite{vireo172} are three large image datasets designed for recipe retrieval and UNICT-FD889~\cite{farinella2014benchmark} and FooDD~\cite{pouladzadeh2015foodd} are designed for food classification. All of these datasets have no food location (i.e., bounding boxes or pixels corresponding to foods) information.
UEC-100~\cite{Matsuda:2012ab} is collected from popular foods in Japan. Similarly, UEC-256~\cite{kawano2014automatic}, the expansion of UEC-100, focus on asian foods only. Although both datasets have food location and category information, only small portion of them are multiple food images.
In UNIMIB2015~\cite{Ciocca2015} and UNIMIB2016~\cite{cioccaJBHI}, food images are captured under controlled environment using the same canteen tray and plates.
The Mixed-Dish~\cite{mixed_dish} is a recently published dataset containing $164$ different asian foods in Singapore restaurants.
It is worth mentioning that all the category names in the datasets we've mentioned above are defined by their authors. Therefore, it is challenging to establish the associations between those semantic food categories and standard food nutrition database such as the FNDDS~\cite{fndds2018}.

\begin{table*}[ht]
\centering
\caption{Food Datasets Used in the Literature}
  \label{tab:dataset_compare}
\begin{threeparttable}[t]
  \begin{tabular}{|c|S[table-format = 8.0]|S[table-format = 4.0]|c|c|c|c|}
    \hline
    \multicolumn{1}{|c|} {Dataset Name}& {\# of}& {\# of}& {Food Type}& {Study Type}& {Annotation}& {\% of }\\
    \multicolumn{1}{|c|} {~}& {Images}& {Cat.}& {~}& {~}& {~}& {Multi-food }\\
    \hline
    PFID~\cite{chen2009} & 4,545&101&Fast Food &Controlled &Label &$-$\\
    Chen~\cite{Chen:2012aa} & 5,000&50&$-$&Free-living &Label &$-$ \\
    Food-101~\cite{bossard14} & 101,000& 101&$-$&Free-living&Label & $-$\\
    UNICT-FD889~\cite{farinella2014benchmark} & 3,583& 889&$-$&Free-living&Label & $-$\\
    UPMC-101~\cite{wang2015recipe} & 90,840& 101&$-$&Free-living&Label & $-$\\
    FooDD~\cite{pouladzadeh2015foodd} & 3,000& 23&$-$&Free-living &Label & $-$\\
    Menu-Match~\cite{beijbom2015menu} & 646& 41&Restaurants & Free-living & Label & $-$\\
    Recipe1M~\cite{recipe1m} & 887,706 & 1,047&$-$ & Free-living & Label & $-$\\
    Recipe1M+~\cite{recipe1m+} & 13,735,679 & 1,047&$-$ & Free-living & Label & $-$\\
    Vireo-172~\cite{vireo172} & 110,241 & 172& Chinese & Free-living & Label & $-$\\   
    \hline
    UEC-100~\cite{Matsuda:2012ab} & 12,740& 100&Japanese &Free-living &BBox &$9.2\%$\\
    UEC-256~\cite{kawano2014automatic} & 28,897& 256&Asian &Free-living &BBox &$6.4\%$\\
    UNIMIB2015~\cite{Ciocca2015} & 2,000& 15&$-$&Controlled & Poly & $100\%$\\
    UNIMIB2016~\cite{cioccaJBHI} & 1,027& 73&$-$&Controlled & Poly & $100\%$\\
    Mixed-Dish~\cite{mixed_dish} & 9,246& 164& Asian&Free-living & BBox & $100\%$\\
    VFN\tnote{1} & 14,991& 82&American &Free-living &BBox& $26.1\%$\\
    \hline
  \end{tabular}
    \begin{tablenotes}
    \item[1] \url{https://lorenz.ecn.purdue.edu/~vfn/}.
    \end{tablenotes}
\end{threeparttable}
\end{table*}

\section{VIPER-FoodNet (VFN) Dataset}\label{Dataset}
\subsection{Food Categories}
What We Eat In America (WWEIA)~\cite{wweia} provides two days of 24-hour dietary recall data which are collected through an initial in-person interview and a follow-up phone interview.
It also shows the intake frequency of each food category during two-day 24-hour recall interviews.
We selected food categories with high intake frequency to create a food image dataset that represents the most frequently consumed foods in the United States. These food categories have associated food codes created by the United States Department of Agriculture (USDA), which can be used to retrieve nutrient information through standard food nutrition database such as the FNDDS~\cite{fndds2018}. 
We selected 82 food categories from the WWEIA database, and used them as food categories in our VFN dataset.

\subsection{Semi-Automatic Food Image Collection and Annotation}
Collecting food images with proper annotations in a systematic way can be very time-consuming and tedious using existing tools (e.g. Amazon Mechanical Turk~\cite{buhrmester2016amazon}).
Therefore, we implemented a semi-automatic data collection system to efficiently collect large sets of relevant online food images~\cite{fang_2018_ssiai}.

Online sharing of food images has gained popularity in recent years on websites such as Facebook, Flickr, Instagram for social networking and Yelp, Pinterest for product review and recommendation. There are also websites dedicated to the sharing of food images, such as yummly\footnote{\url{https://www.yummly.com/}} and foodgawker\footnote{\url{https://foodgawker.com/}}. There are hundred-thousands of food images uploaded by smartphone users to these websites. Online food images also provide valuable contextual information which is not directly produced by the visual appearance of food in the image, such as the users' dietary patterns and food combinations~\cite{wang2018context}.
To quickly collect a large number of online food images, we implemented a web crawler to automatically search on the Google Image website based on selected food labels and download the retrieved images according to the relevant ranking on the Google Image.

However, some of these retrieved images are considered as noisy images which do not contain relevant foods.
Following the method described in~\cite{Shao2019}, we trained a Faster R-CNN \cite{ren2015faster} for food region detection to remove non-food images.
Food images passed automatic noisy image removal step are assigned for further confirmation and food item localization in the online crowdsourcing tool. 
The crowds are asked to draw a bounding box around each food item in the image, and select the food category associated with each bounding box.
Using this semi-automatic crowdsourcing tool, We created the VIPER-FoodNet (VFN) dataset which contains 82 food categories, 14,991 online food images and 22,423 bounding boxes.

\section{Food Recognition System}\label{recognition}
\begin{figure*}[ht]
\begin{center}
  \includegraphics[width=1.0\textwidth]{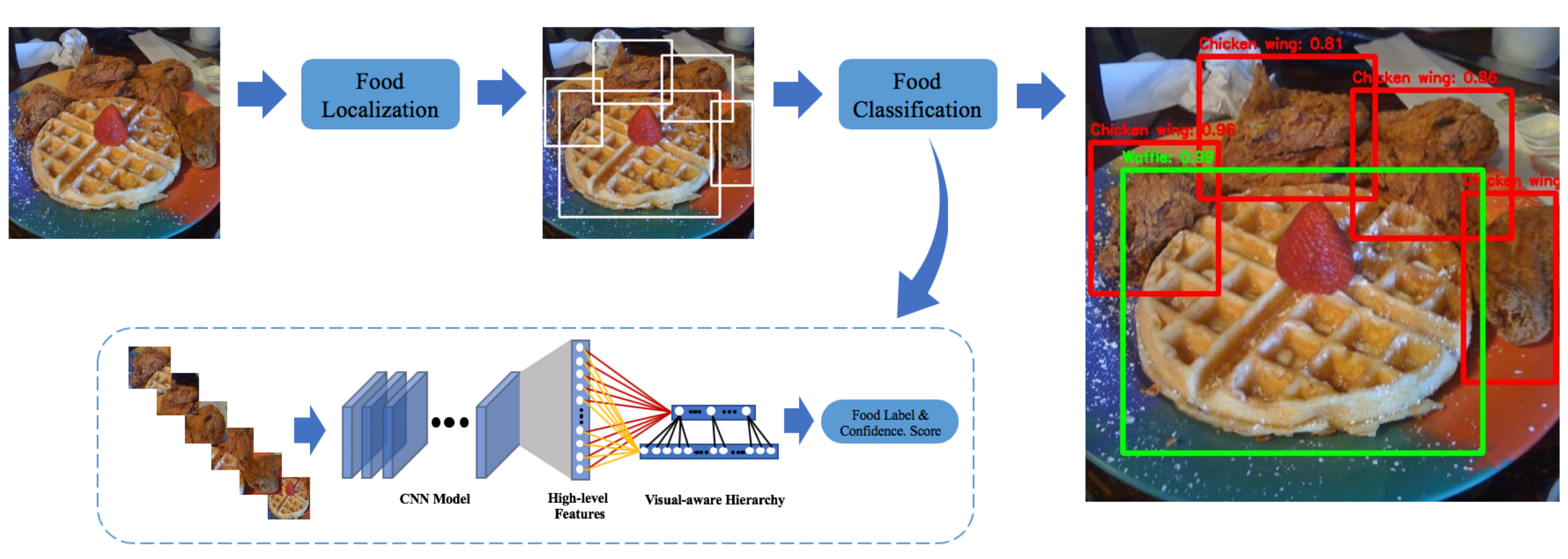}
  \caption{Our proposed food recognition system. The food localization step, trained on Faster R-CNN, selects regions in an input image that contain food item. The selected regions, resized to $224 \times 224$, are fed into the food classification system. The visual-aware hierarchical structure is built based on the features extracted from a CNN model. Food label, cluster label and associated confidence scores can then be predicted for each selected region in a multi-task manner. The final output includes the bounding box and food label for each food in the input image.} 
  \label{fig:1}
\end{center}
\end{figure*}

Dietary assessment, which collects what an individual eats during the course of a day, can be time-consuming, tedious, and error-prone using traditional methods that rely solely on human memories and/or recordings. To improve the efficiency and accuracy of dietary assessment, we propose an automated image-based food recognition system shown in Fig.\ref{fig:1}. Our system consists of two parts: food localization and food classification. The goal of food localization is to locate each individual food region in an image with a bounding box. Pixels within the bounding box are assumed to represent a single food, which is the input to the food classification. Food localization serves as a pre-processing step since it is common for food images in real-life to contain multiple foods. 
In addition, food localization can remove non-food background in the image to improve the classification performance. Without this step, the food classification task is much more challenging. Different from existing CNN based food classification systems, the visual-aware hierarchical structure is embedded in a multi-task CNN model in our food classification method. Our method first clusters visually similar food categories based on learned features, and then leverages the hierarchical structure and a multi-task CNN model to embed visual relations between different food categories and improve the classification performance. We describe the details of each component of our system in this section. 

\subsection{Food Localization}\label{Localization}
We are interested in finding regions in a food image that contain foods, in particular, each region should contain just one food. Deep learning based methods such as Faster R-CNN~\cite{ren2015faster} and YOLO~\cite{redmon2016you} have shown success in many computer vision applications. 
Most mainstream CNN architectures such as VGG~\cite{vgg} and ResNet~\cite{resnet} can be used as the backbone structure for these methods. In our method, we use Faster R-CNN with VGG-16 as the backbone for food localization. 

Faster R-CNN proposes potential regions that may contain the object with bounding boxes. It also assigns a confidence score to each bounding box. In our system, we call this confidence score the ``foodness" score since it is the confidence score of food regions in the image. A high ``foodness" score indicates a high likelihood that the region contains food. In our implementation, we estimate the ``foodness" score threshold based on validation dataset's performance. Regions with ``foodness" score above that threshold is fed into the food classification step.  
To train the Faster R-CNN, all different food categories in the dataset is treated as one category, i.e., food. For each food dataset, we selected $70\%$ of the images as training data, the rest $10\%$ is used for validation to determine the optimal hyper-parameter in Section~\ref{sec:localization_exp}, and $20\%$ for testing the performance of food localization. 
The Faster R-CNN method consists of a Region Proposal Network (RPN) and a Classifier. A Non-Maximum Suppression (NMS) threshold is also selected to remove redundant regions.

\subsubsection{Region Proposal Network (RPN)}
This network is used to suggest foreground object regions in the image. Before RPN, feature map is generated based on the last convolution layer. The RPN generates 9 different sized anchor boxes by sliding a small network over the feature map. Each anchor returns the foreground object confidence score and a set of bounding box coordinates. After Non-Maximum Suppression (NMS) in Section~\ref{NMS}, features inside the anchor boxes are used by a classifier to determine whether it contains food or not.

\subsubsection{Classifier}
Since different anchor boxes have different dimensions, the Region of Interest (RoI) pooling~\cite{girshick2015fast} is used to create the fixed size feature maps. The fully-connected layer, the classifier, will predict the generic labels and assign the confidence score for each selected region. The confidence score ranges from 0 to 1, and is the probability of the predicted label for each region. For example, if our system assigns 0.65 to a region of the input image, that means our system believes this region has $65\%$ probability of containing food. In our case, this is a binary classification task, which identifies whether a region contains food or not. We label this confidence score as the ``foodness" score. Ideally, the classifier should assign high ``foodness" score to food regions and low score to non-food regions. 

\subsubsection{Non-Maximum Suppression (NMS)}\label{NMS}
\begin{figure}[ht]
\begin{center}
  \includegraphics[height=1\textwidth,angle =90]{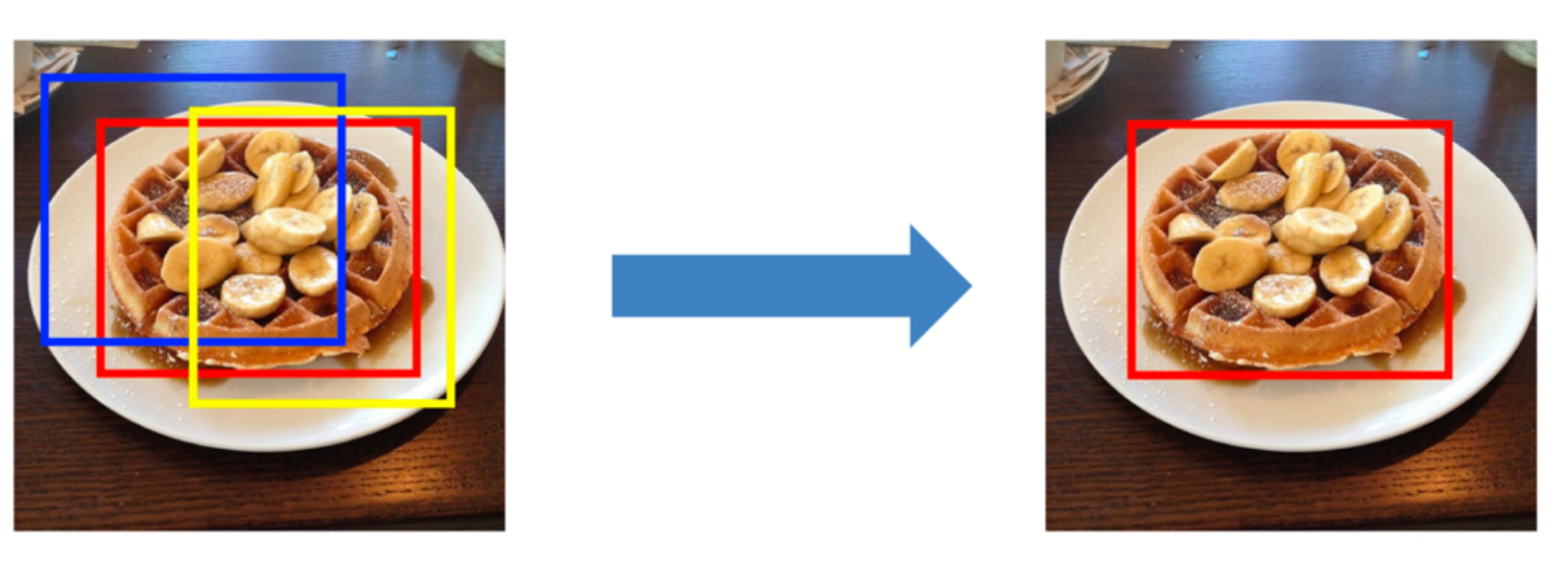}
  \caption{An example of applying Non-Maximum Suppression (NMS) to proposed food regions. In the left image, all three regions have high confidence scores and the IoU is larger than 0.7. In the right image, NMS selects the region with the highest confidence score.}
  \label{fig:2}
\end{center}
\end{figure}

The RPN may propose regions that are highly overlapped. An example is shown in Figure~\ref{fig:2}. We adopt Non-Maximum Suppression (NMS) to select the bounding box with the highest ``foodness" score and remove other bounding boxes with significant overlap. Intersection Over Union (IoU) is used to measure how significant the overlap is. As shown in Equation~\ref{eq:1}, B1 and B2 correspond to two bounding boxes. Following~\cite{ren2015faster}, we set the IoU threshold to 0.7. In our implementation, if there are several bounding boxes with IoU value larger than 0.7, we retain the bounding box with the largest ``foodness" score.

\begin{equation} \label{eq:1}
    IoU = \frac{B_{1} \cap B_{2}}{B_{1} \cup B_{2}}
\end{equation}

\subsection{Food Classification}
Convolutional Neural Networks (CNNs) have been widely used in image classification applications. Instead of training a flat CNN structure using food image datasets, we propose a hierarchical structure to further improve the accuracy and to make better mistakes. For vanilla CNN-based image classification method, the feature map will densely connect to the top layer, whose length equal to the number of categories. Each ground-truth label will be encoded in the one-hot sequence for computing cross-entropy loss. One-hot representation of N class will have N binary bits with one single high ($1$) bit and all the others low ($0$). One-hot encoding makes the $L_{p}$ distances between different category label equally. In other words, the distance between label ``Bagel" and ``Bread" is the same as that between ``Bagel" and ``Pork chop" in encoding space. To reduce the information loss in this encoding process, we proposed to embed the hierarchical structure to represent visual relationships between food categories. In~\cite{wu2016learning}, a pre-defined semantic hierarchical tree is proposed, which contains food clusters with semantic similar food categories. Results showed that using the hierarchical structure can improve the accuracy of classification and make better mistakes. However, the hierarchy is defined manually, and customized for Food-101~\cite{bossard14} and 5-Chain~\cite{wu2016learning}.  Most existing food image datasets~\cite{kawano2014automatic,Matsuda:2012ab,bossard14,upmc} contain different food categories. As a result, the semantic-aware hierarchical structure need to be rebuilt for different datasets. It is worth mentioning that same food may have different names, e.g. courgette and zucchini, and same word may refer to different foods, e.g. muffin in England and in America are different. It would be very time-consuming to define a specific semantic-aware hierarchical structure for each dataset. In this paper, we propose a method to cluster visually similar food categories and automatically generate a hierarchical structure to improve CNN based food classification.

\subsubsection{Food Similarity Measure}\label{similarity}
Many existing methods can cluster similar categories based on semantics. However, food categories with high semantic relations do not always share similar visual features. In addition, food categories vary in different regions of the world. The same food may have different names in different cuisine. Manually record each food category and compute semantic relation is expensive and not feasible for large datasets. Therefore, semantics structure is hard to build and the semantics correlation may mislead visual-feature based training process if semantic similar categories have distinctive visual appearances. CNNs are commons used in image classification to extract visual features. These visual features can be used to identify correlations between different food categories. Based on the feature map of the convolutional layer, we can compute the visual similarity between food categories and cluster similar categories automatically.

In our system, we choose DenseNet-121~\cite{huang2017densely} model as our visual feature extractor. It consists of convolutional layers and full-connected layers. Convolutional layers are used to extract features from input images. Fully-connected layers are used to perform the classification task based on the features generated by convolutional layers. The output of the last convolution layer is treated as the feature map for each food image. Our feature map for each input image is a $1\times1024$ space vector and represents one data point in the $1,024$ dimensional feature space. We used the pre-trained model on the ImageNet~\cite{ilsvrc_15} dataset and use a small learning rate to fine-tuning the pre-trained model to reduce training time. The learning rate is set to 0.0001. The model is flat trained on the food dataset and we used the cross-entropy loss as the loss function.

    \begin{equation} \label{eq:5}
    Cross~Entropy~Loss=-\sum\nolimits_{i=1}^{N}y_{i}log(p_{i})
    \end{equation}

\begin{figure}[ht]
\centering
\subfloat[\label{fig:3a}]{\includegraphics[width=0.32\textwidth]{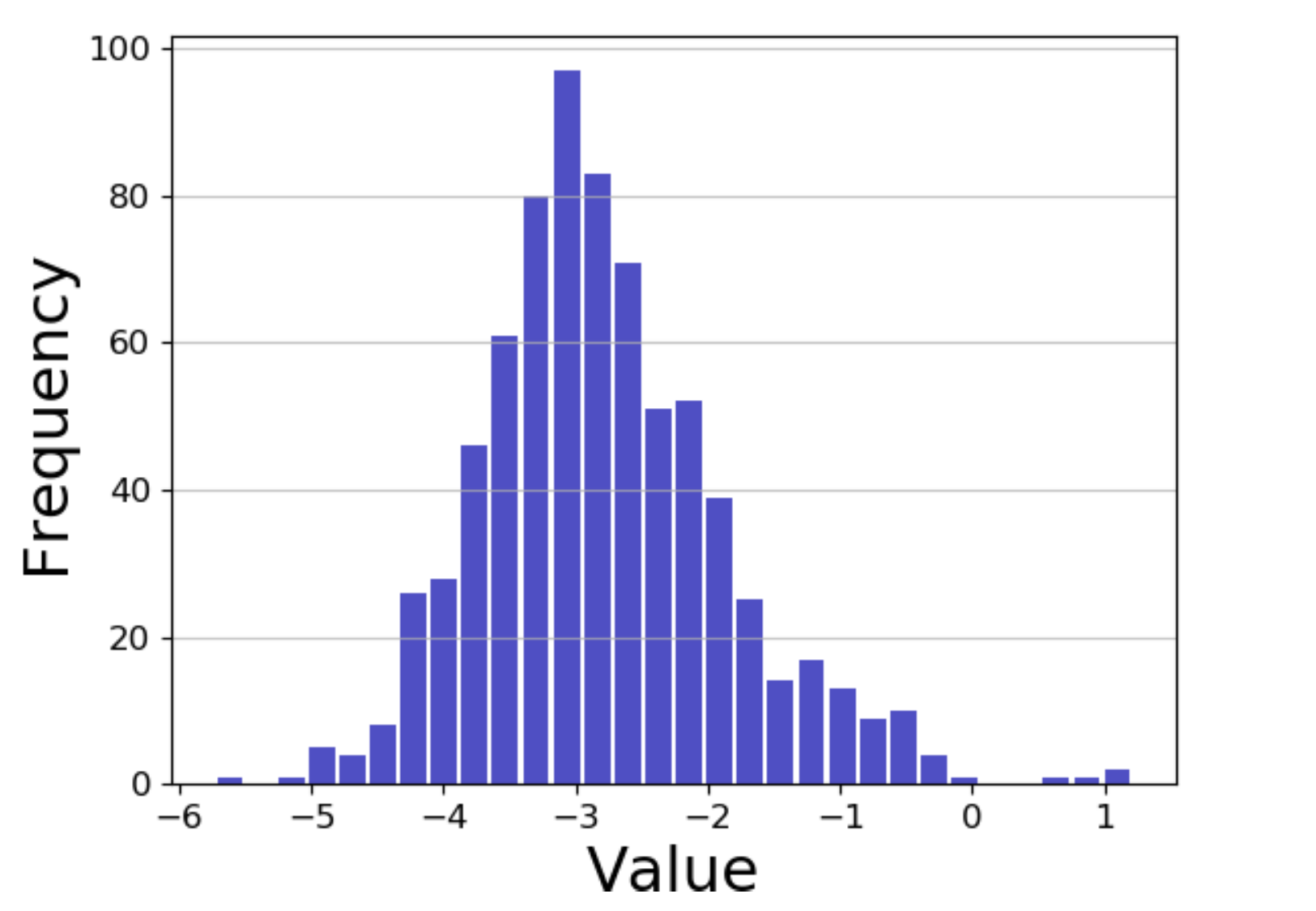}}
\subfloat[\label{fig:3b}]{\includegraphics[width=0.32\textwidth]{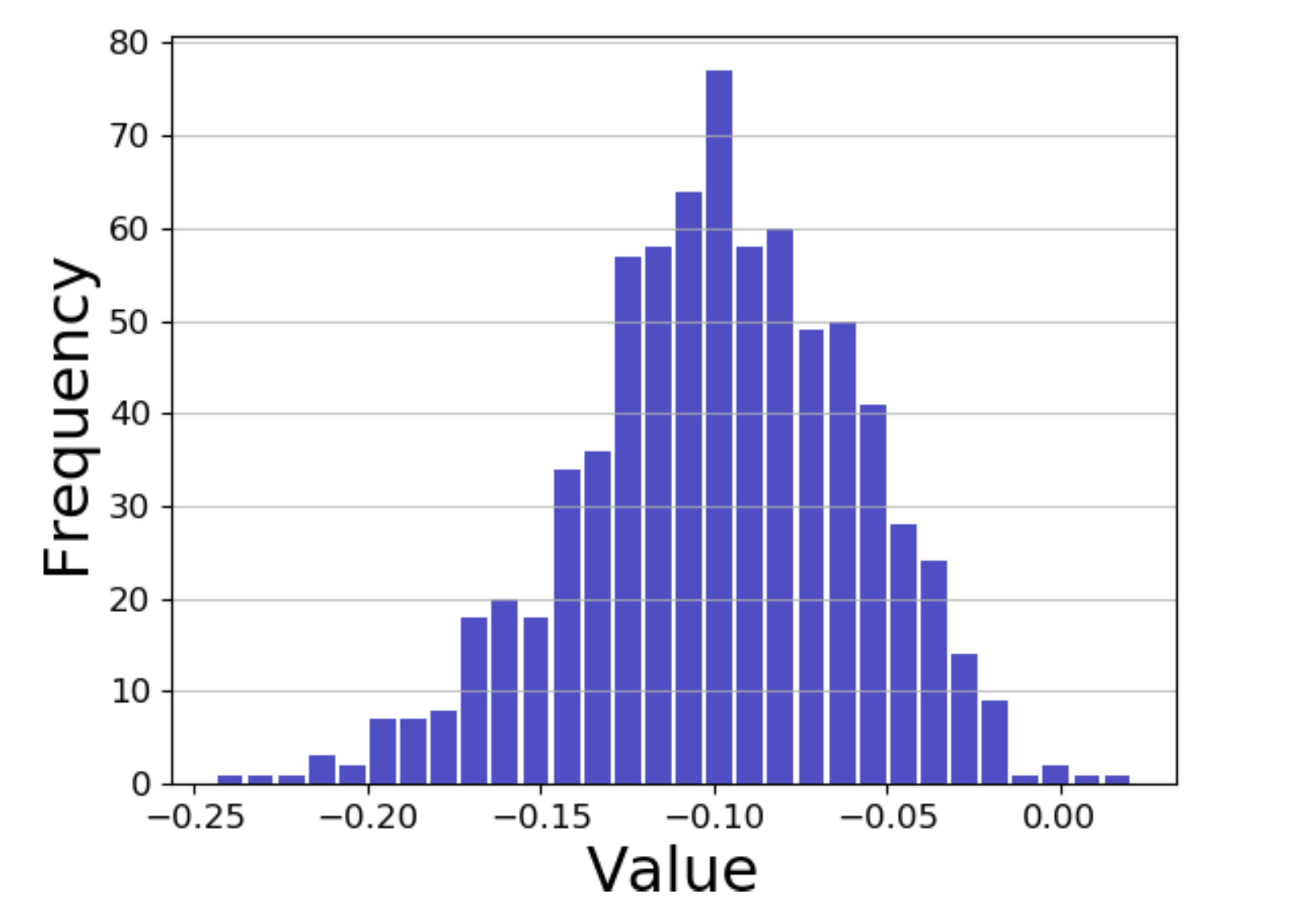}}
\subfloat[\label{fig:3c}]{\includegraphics[width=0.32\textwidth]{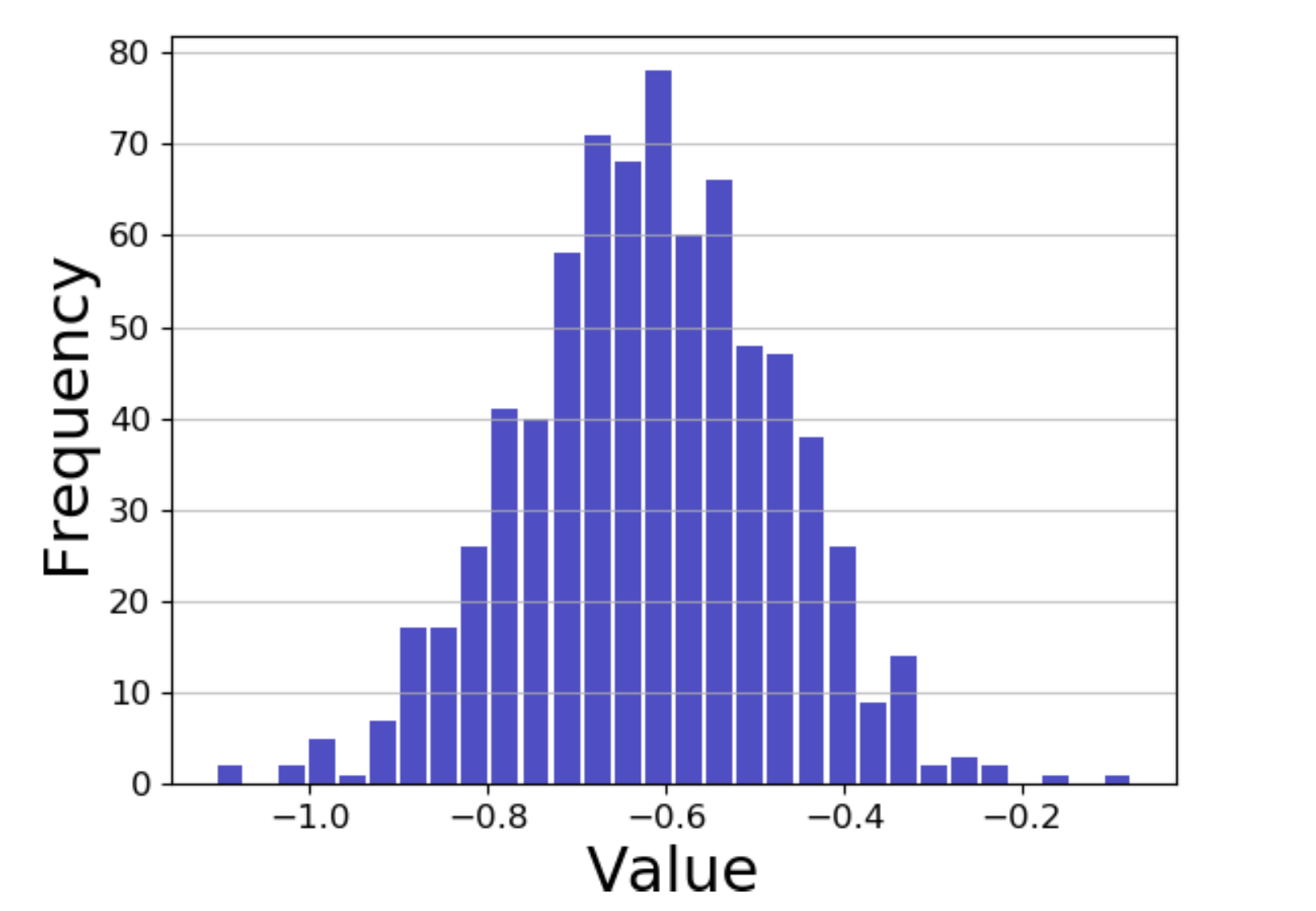}}
\caption{{The feature map of the input image is a $1 \times 1024$ vector represents 1024 feature dimensions. Different food categories have different feature distributions. This figure shows histograms of 3 of 1024 feature dimensions for all training images of apple pie. All three features exhibit Gaussian-like distribution.}}
\label{fig:feature}
\end{figure}

As shown in Equation~\ref{eq:5}, N is the total number of classes, $y_{i}$ and $p_{i}$ correspond to the $i^{th}$ element in \textbf{y} and \textbf{p}, whose length are N. \textbf{y}, the ground truth label, is encoded as one-hot sequence, and \textbf{p} is the confidence scores of each category predicted by networks.
Once the loss is converged, the model then has the ability to extract meaningful visual features for food classification. 
If the model is well trained, as show in Figure~\ref{fig:feature}, each dimension of the feature map of one food category should have a Gaussian-like distribution. Therefore, we can generate a 1D Gaussian Probability Density Function to fit the distribution and compute their Overlap Coefficient (OVL)~\cite{OVL}.

\begin{figure}[ht]
    \begin{center}
      \includegraphics[width=1\textwidth]{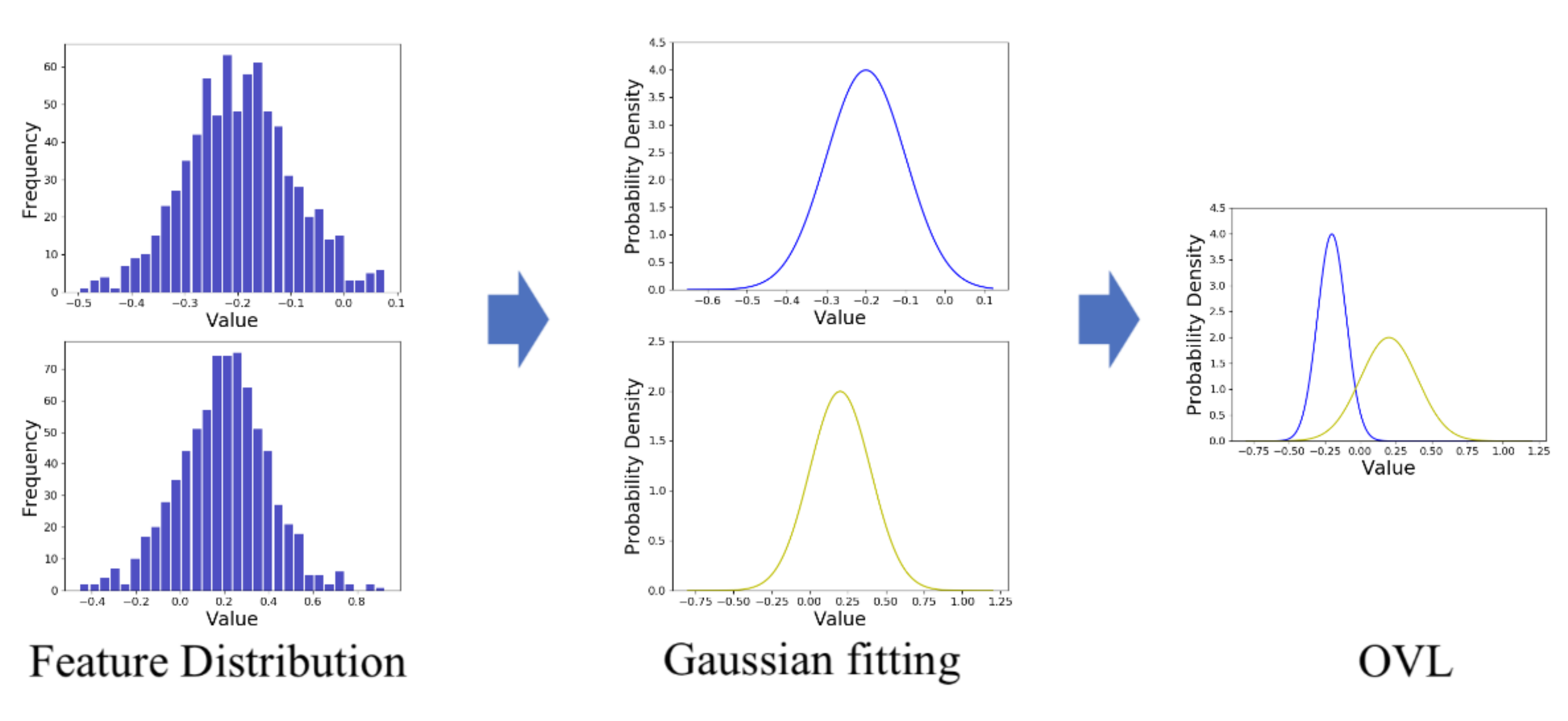}
      \caption{The left two plots are distributions of two different food categories in one specific high-level feature. We fit Gaussian Probability Density Function (PDF) over this distribution. The overlap coefficient (OVL) is the overlapped area between the two PDFs.}
      \label{fig:4}
    \end{center}
    \end{figure}

\begin{figure*}[ht]
    \centering
    \subfloat[]{\includegraphics[height=0.49\textwidth,angle =90]{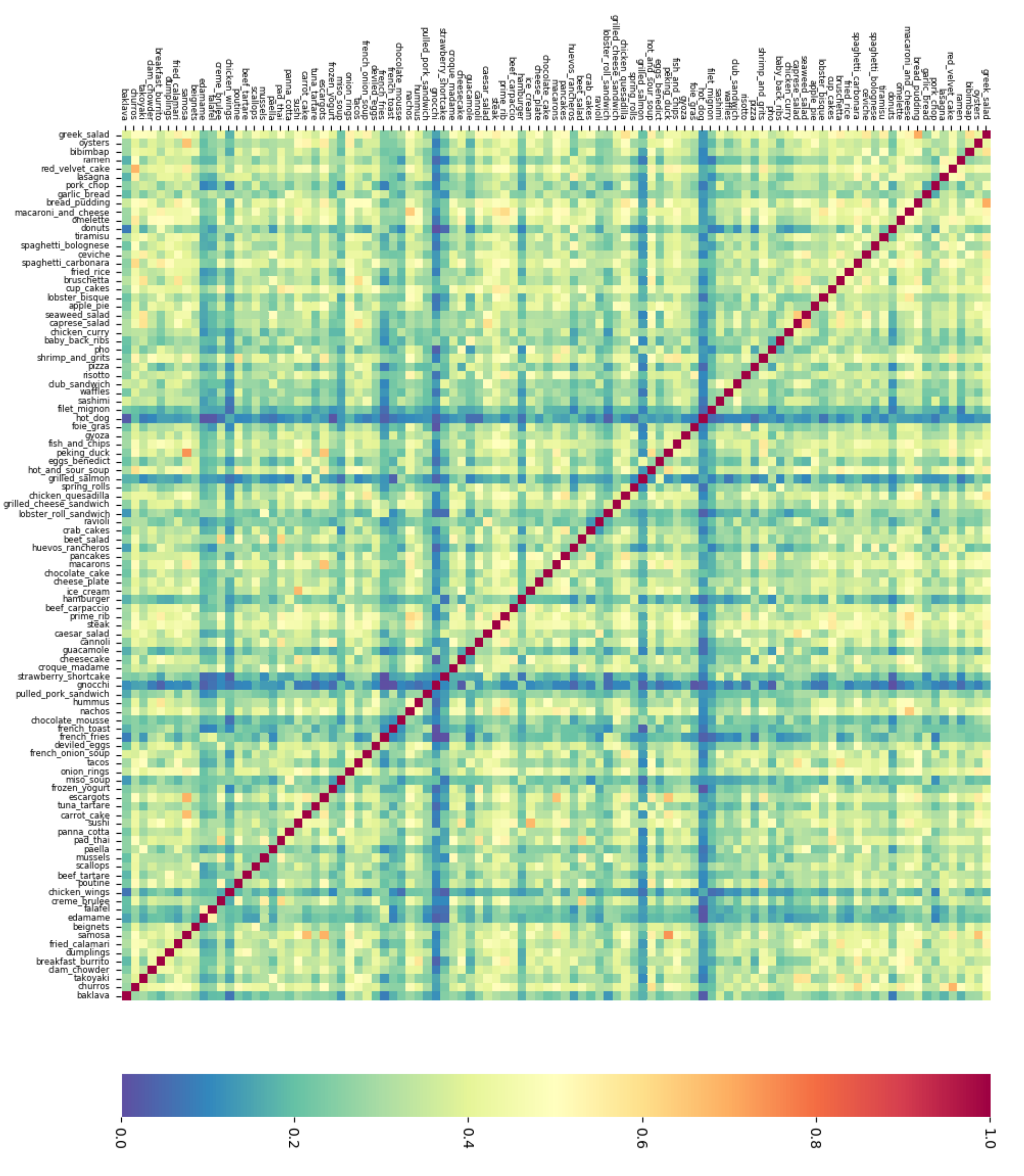}
      \label{fig:5a}}
      \subfloat[]{\raisebox{3.3ex}{\includegraphics[width=0.49\textwidth]{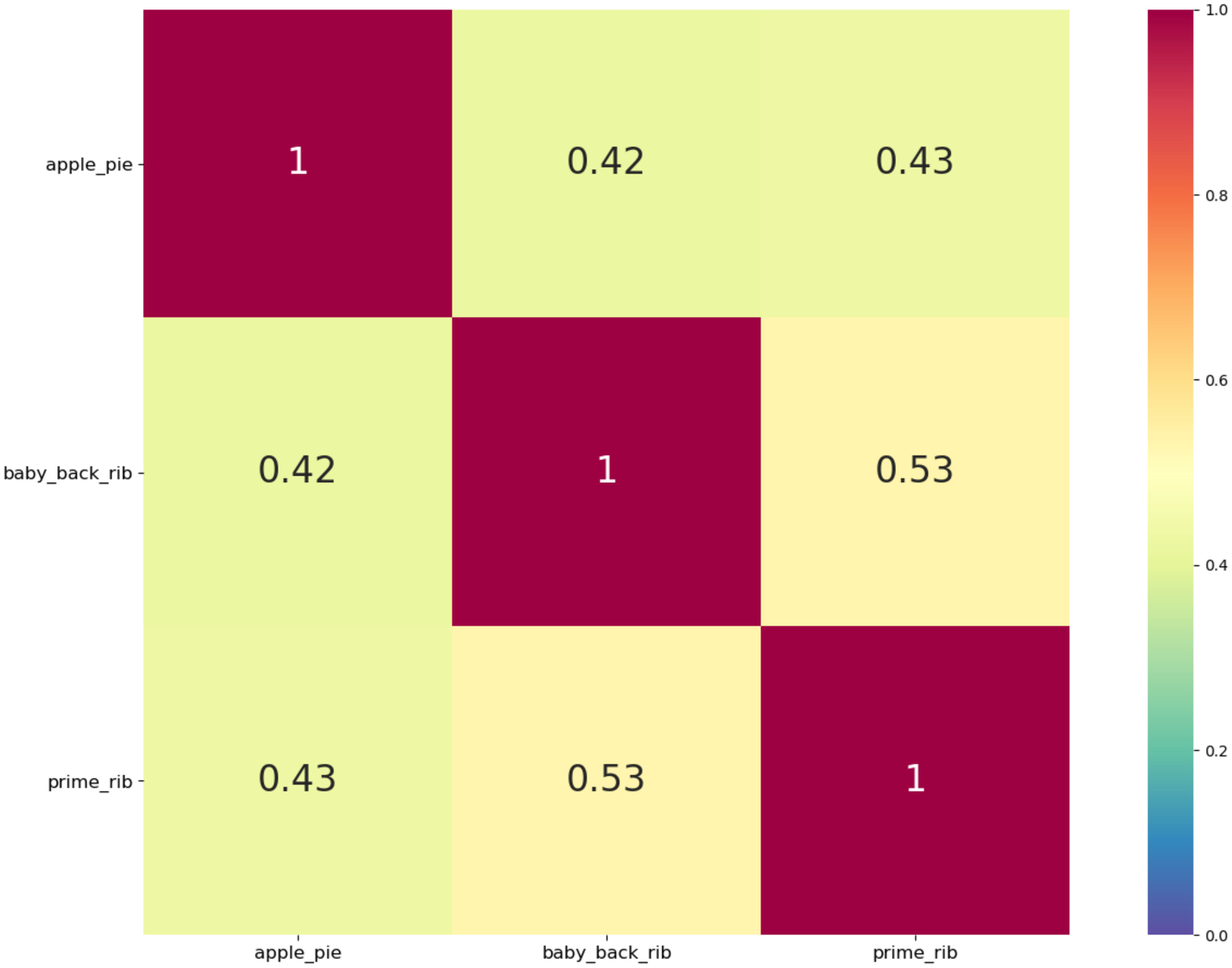}}
      \label{fig:5b}}
          \caption{Similarity matrix: (a) shows the entire similarity matrix of the Food-101 dataset. In (b), we select three categories: Apple pie, Baby back rib, and Prime rib. Apple pie is quite different from the other two categories. Prime rib and Baby back rib are visually similar, which indicated by higher similarity scores.}
        \label{fig:5}
    \end{figure*}

As shown in Figure~\ref{fig:4}, the OVL refers to the area under two probability density functions. It is a measure of the agreement between two distributions. Therefore, if two food categories have high OVL in one dimension of the feature map, that means both food categories are similar with respect to this feature dimension. As a result, we compute the OVLs in all 1,024 dimensions and normalize them to generate the similarity matrix. Figure~\ref{fig:5a} shows the similarity matrix of the Food-101 dataset food categories. We select three categories, as shown in Figure~\ref{fig:5b}, to show an example of the similarity measure between different food categories. For instance, baby back rib and prime rib show higher similarity (0.53) compared to apple pie (0.42). 

As shown in Equation~\ref{eq:5}, cross-entropy uses one-hot sequence to encode the ground truth label. Due to the nature of one-hot encoding, each pair of labels has the same $L^{p}$ distance and there is no visual relations embedded.
However, visually similar food categories obtain higher similarity score in our similarity matrix in Figure~\ref{fig:5}. We propose to cluster those similar categories and build a hierarchy to express their relations in the explicit way.

\subsubsection{Food Clustering and Hierarchical Structure}\label{hierarchy}
Hierarchical structure can represent the relations between food categories. In addition, it is the key to build the multi-task model for further improving the performance of the classification task. To generate the hierarchical structure for our system, we need to cluster similar food categories first. Based on the similarity matrix calculated above, many clustering methods can be applied. Although the existing method, such as K-means, can partition a dataset into K clusters efficiently and find the centroids of the K clusters accurately, it requires pre-defined number of clusters, k. Unlike K-means, Affinity propagation (AP) \cite{frey2007clustering} does not need to know the number of clusters a priori, instead, it can determine the optimal cluster number. In our implementation, we use AP to cluster the similar food category and generate a 2-level hierarchical structure. AP treats all food categories as candidates and selects m candidates as exemplars to represent m clusters separately. It iteratively refines the selection until it reaches the optimal solution. The input of AP is the similarity matrix (s) calculated in Section~\ref{similarity}. We first define two matrices ``Responsibility" (r) and ``Availability" (a). Initially, both matrices are set to zero. Both matrices are then updated alternately as shown in Equation~\ref{eq:6} and Equation~\ref{eq:7}.

    \begin{equation} \label{eq:6}
    r(i,k) = s(i,k)-\textrm{max}_{k^{'} \neq k}(a(i,k^{'})+s(i,k^{'}))
    \end{equation}  
    
    \begin{equation} \label{eq:7}
    \begin{aligned}
    a(i,k)_{i\neq k} = &\textrm{min} \Big( 0,r(k,k)+\sum_{i^{'}\notin \{i,k\} }\textrm{max}(0,r(i^{'},k)) \Big)\\
    &a(k,k) = \sum_{i^{'}\neq k }\textrm{max}(0,r(i^{'},k))
    \end{aligned}
    \end{equation}  
    
The three matrix: s, a, r, have size N x N, where N refers the total number of categories. In Equation~\ref{eq:6}, $r(i,k)$ quantify how well-suited $k^{th}$ category is to be the exemplar for $i^{th}$ category, relative to other candidate exemplars. It is the information propagated from $i^{th}$ category to $k^{th}$ category. In Equation~\ref{eq:7}, $a(i,k)$ quantify how appropriate for $i^{th}$ category to pick $k^{th}$ category as its exemplar, which is a feedback propagated from $k^{th}$ category. If $r(i,i)+a(i,i)>0, x_{i}$ is selected as the exemplar. Each following iteration will update the selected exemplars. If the selection does not change for more than 15 iterations, the result is said to reach the optimal solution. Once we formed the stable clusters, we can build a hierarchical structure based on the clustering result. Using the same strategy, similar clusters can be further grouped for generating three or higher-level hierarchy.

\subsubsection{Multi-task Learning for Deep Neural Networks}
\begin{figure}[ht]
    \begin{center}
      \includegraphics[width=1\textwidth]{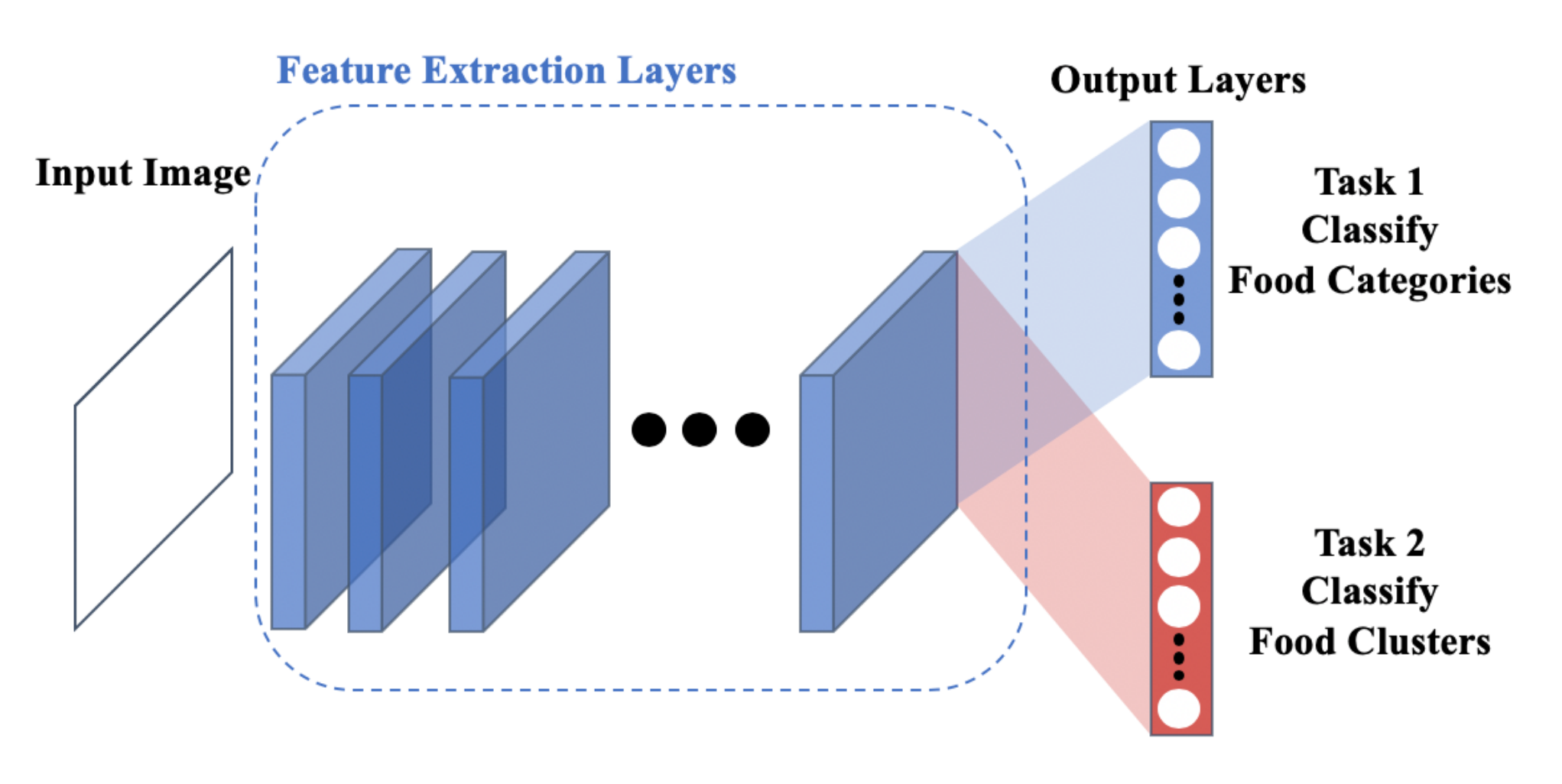}
      \caption{A two-level multi-task CNN architecture: The feature extraction layers for joint feature learning  output the high-level features. Two separated output layers are fully connected to those features for category prediction (Task 1) and cluster prediction (Task 2)}
      \label{fig:6}
    \end{center}
    \end{figure}

Multi-task models are employed for obtaining predictions for several related tasks simultaneously. Suppose the hierarchical structure generated from Section~\ref{hierarchy} has two levels, then there are two related tasks assigned to the multi-task model, i.e. $1^{st}$-level category predication and $2^{nd}$-level cluster prediction. As shown in Figure~\ref{fig:6}, the multi-task deep learning model proposed in this paper contains two parts: feature extraction layers and output layers. 
Unlike the visual feature extractor trained by regular cross-entropy loss as described in Section~\ref{similarity}, the feature extraction layers for joint feature learning are trained by multi-task loss back-propagated from output layers. The number of the output layers is equal to the number of levels in the hierarchy. The nodes in each layer correspond to labels of that level. All the output layers are fully-connected to the high-level features learned from joint feature learning for predictions.

Suppose we have a hierarchical structure $\psi = \{Y^{(t)}\}_{t=1}^{T}$ with T levels, where $\{Y^{(t)}\}$ represent the $t^{th}$ level's label set of the given T-level hierarchical structure. Each node in each level will be assigned a label, e.g. $\{y_{i}^{(1)}\}_{i=1}^{N_{1}}$ represents the original category set and $\{y_{i}^{(2)}\}_{i=1}^{N_{2}}$ represent the label set for the cluster in the second level. Thus the multi-task loss function is formulated as 
    \begin{equation} \label{eq:8}
    L(\textbf{w}) = \sum_{t=1}^T\lambda_t\sum_{i=1}^{N_{t}} -log p(y_i^{(t)}|\textbf{x}_i,\textbf{w}_{0},\textbf{w}^{(t)}) 
    \end{equation}
where $y_i^{(t)} \in Y^{(t)}$ is the corresponding class/cluster label for the $t^{th}$ hierarchical level. $\textbf{w}^{(t)}$ represent the network parameters for the $t^{th}$ output layer, and $\textbf{w}_{0}$ compose the parameters for feature extraction layers. $\lambda_t$ is the hyperparameter that controls the weight of the $t^{th}$ level contribution in the given hierarchical structure.
% the contribution of the fine-grained classification from the t^{th} leaf level of the given hierarchical structure.
During the training process, the weights of the shared feature extraction layers are initialized using the values of the corresponding network pre-trained on ImageNet dataset~\cite{krizhevsky2012imagenet}, while the parameter  $\textbf{w}^{(t)}$ for the $t^{th}$ output layer are learned from scratch. In addition, for the two-level hierarchy proposed in this paper, we set $\lambda_{1}=\lambda_{2}=0.5$ in the multi-task loss.

\section{Experimental Results}\label{Experiment}
\label{sec:result}
Our food recognition system consists of food localization and food classification. In this section, we first evaluated the performance of food localization and food classification separately, and then test the overall performance of the food recognition system. The datasets we used for our experiments include Food-101~\cite{bossard14}, UPMC-101~\cite{upmc}, UEC-100~\cite{Matsuda:2012ab}, UEC-256~\cite{kawano2014automatic} and our proposed VFN datasets.
It is worth mentioning that only UEC-100, UEC-256, and VFN datasets contain bounding box information for food images, which can be used for food localization and food recognition evaluation.
For classification task, our hierarchy-based classification method is tested on Food-101, UPMC-101, which contain only single food images, and cropped single food images from UEC-100, UEC-256, and VFN datasets.
The authors of Food-101 and UPMC-101 established the training and testing sets. We followed their splits and report the standard multi-class accuracy on each testing set.
Following the setting for the recognition task in \cite{Bolaos2016SimultaneousFL}, we applied a random $70\%/10\%/20\%$ split of images for training/validation/testing on each food category for UEC-100, UEC-256, and VFN datasets.
The validation set of each dataset is used to determine the best ``foodness" score threshold to optimize food localization performance.
The cropped images from these datasets are used as training/validation/testing images for classification. Since testing sets are not revealed to both classification and localization, there is no overfitting occurred during the recognition evaluation.
Although UEC-100 is the predecessor of UEC-256, which means it is a subset of UEC-256, many previous works~\cite{CNNfood,yanai2015food,Hassannejad2016} were evaluated on UEC-100, and the expanded 156 categories made UEC-256 quite different from UEC-100. Therefore, we evaluated our system on both datasets for comparison. 

\subsection{Food Localization}\label{sec:localization_exp}
As mentioned in Section~\ref{Localization}, food localization is trained on the training set of each food image datasets, i.e., UEC-100, UEC-256, and VFN datasets. The ``foodness" score threshold is determined by the validation set. Localization performance on the testing sets of each dataset is reported in Table~\ref{tab:1}.

\begin{figure*}[ht]
    \centering
    \subfloat[]{\includegraphics[width=0.32\textwidth]{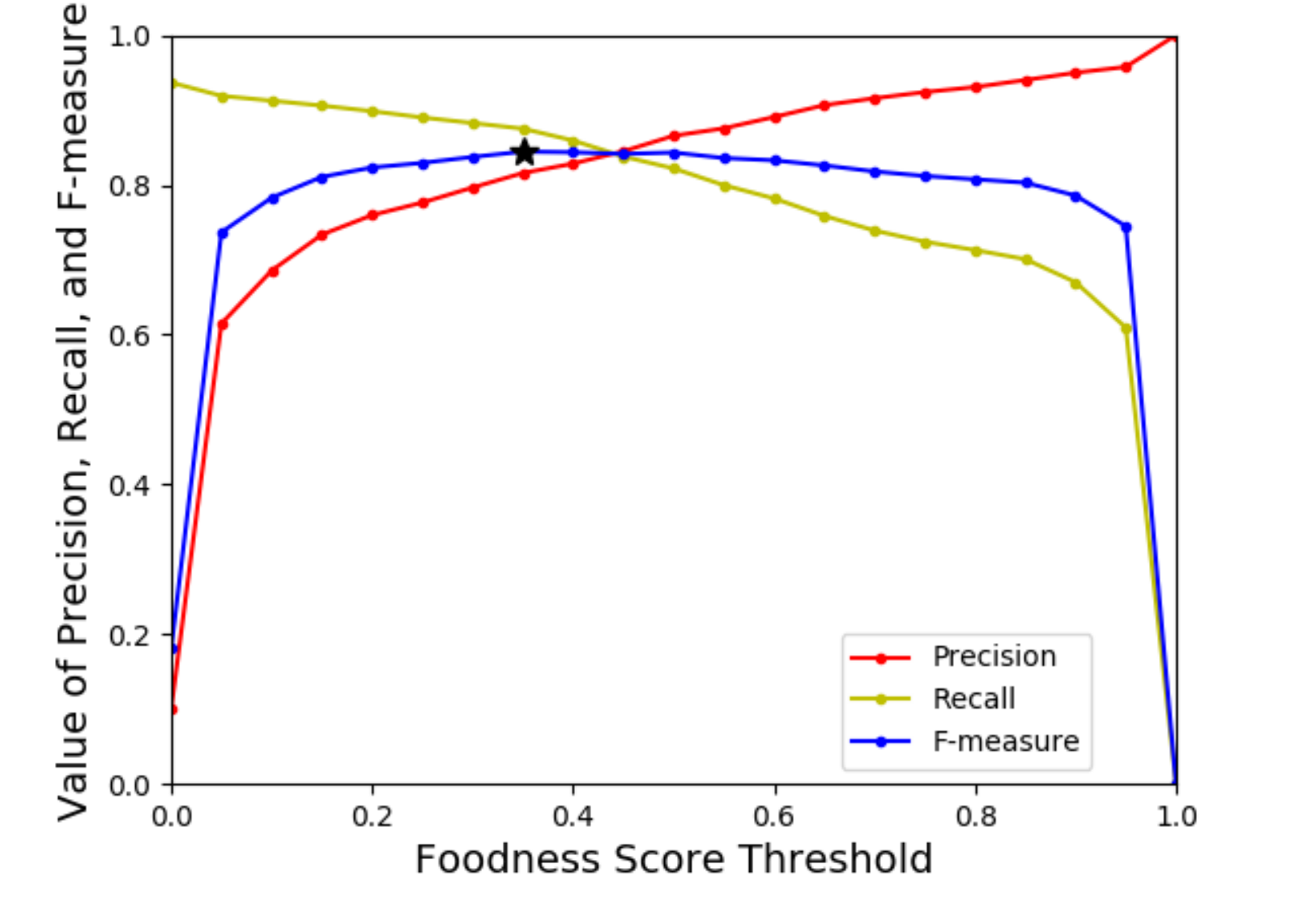}
      \label{fig:7a}}
    \subfloat[]{\includegraphics[width=0.32\textwidth]{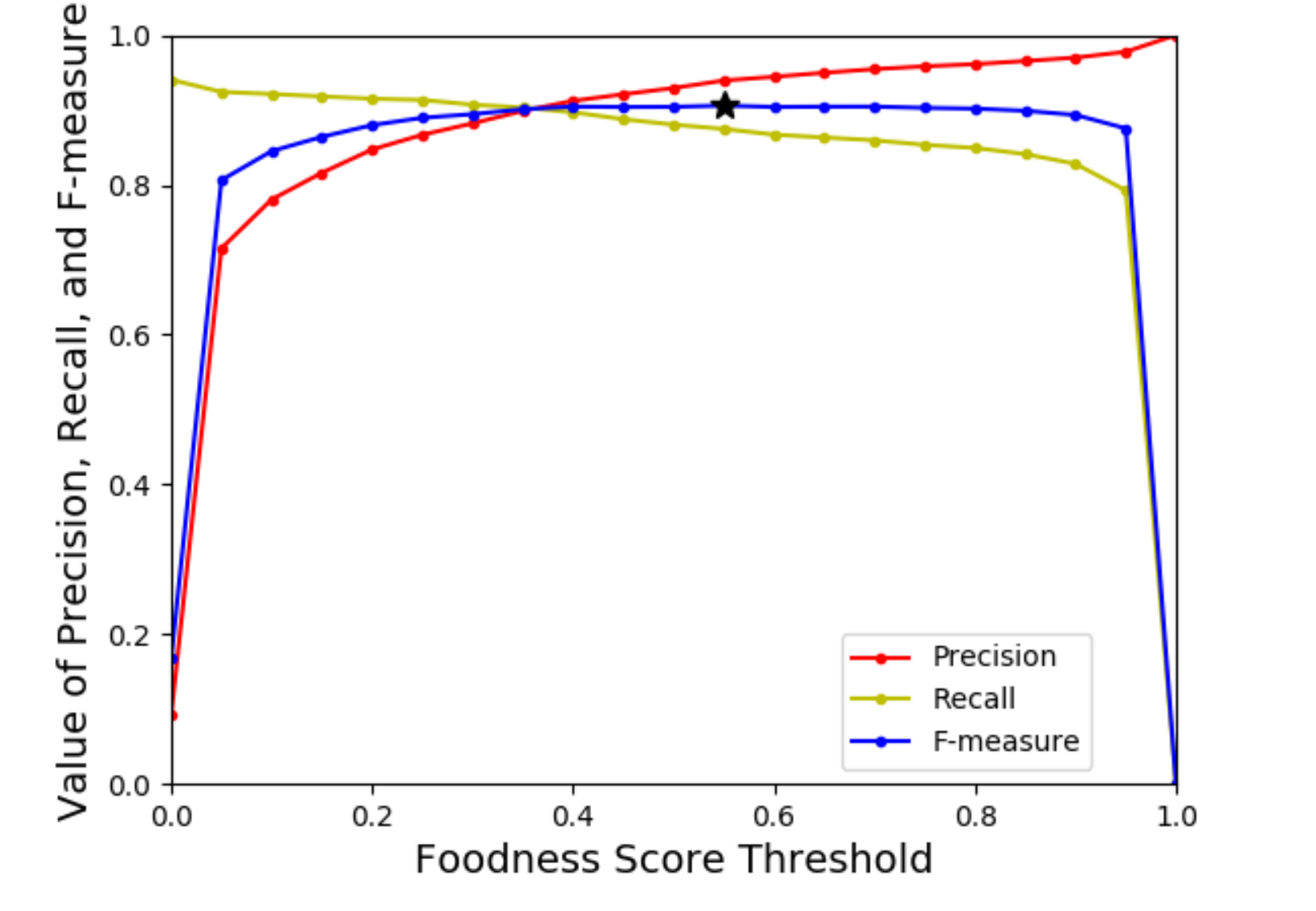}
      \label{fig:7b}}
    \subfloat[]{\includegraphics[width=0.33\textwidth]{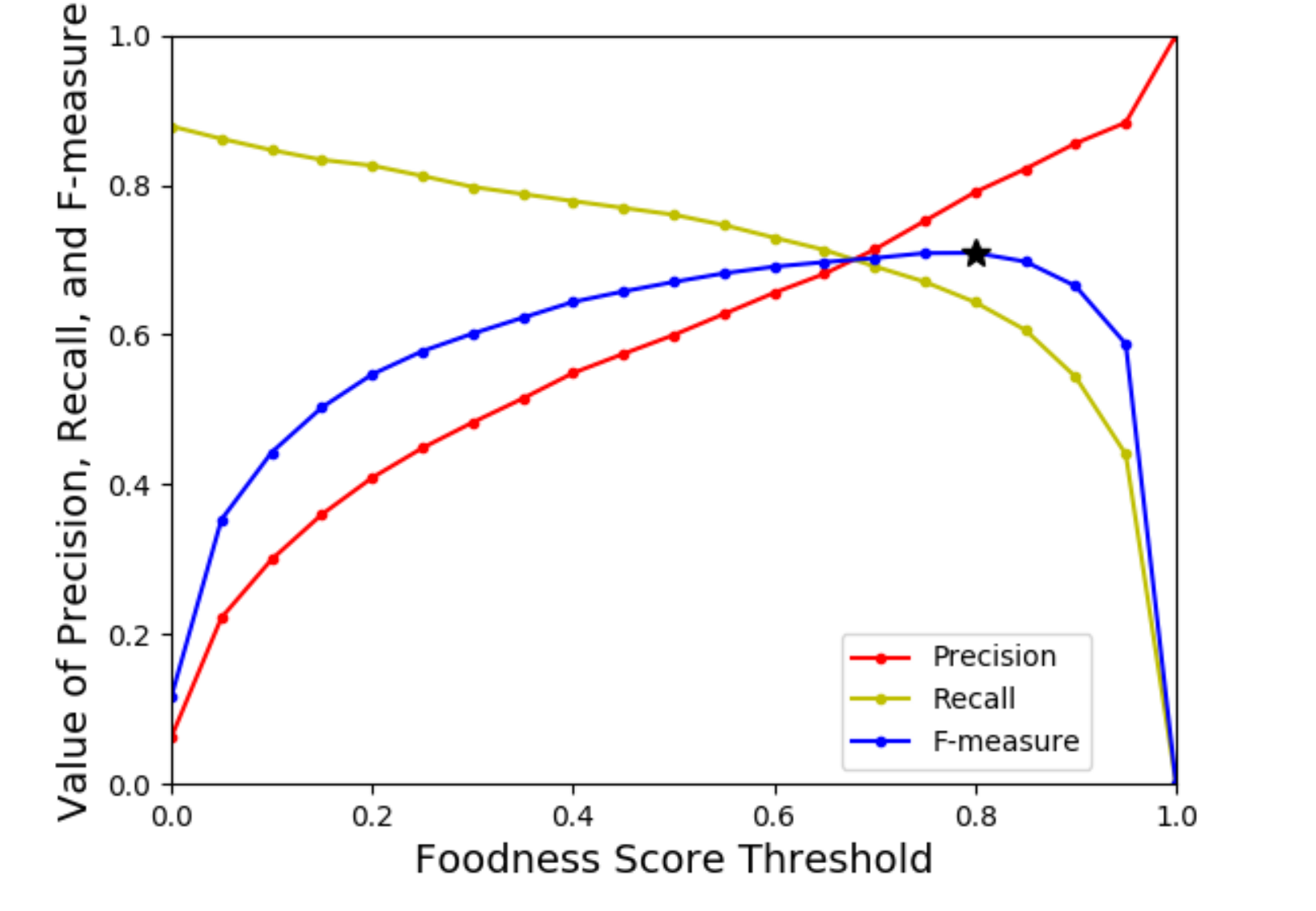}
      \label{fig:7c}}
          \caption{Precisions, Recall, and F-Measure corresponding to different ``foodness" score threshold: (a) UEC-100 dataset has the highest F-Measure when ``foodness" score threshold =  $0.35$. (b) UEC-256 dataset has the highest F-Measure when ``foodness" score threshold = $0.55$. (c) VFN dataset has the highest F-Measure when ``foodness" score threshold = $0.80$.}
        \label{fig:7}
    \end{figure*}

Precision and recall are the most common performance metrics used for localization evaluation. We define four related terminologies: True Positive (TP) means a food region is correctly detected; False Positive (FP) means a non-food region is incorrectly detected as a food region; True Negative (TN) means a non-food region is correctly detected; False Negative (FN) means a food region is incorrectly detected as a non-food region. We define a region is correctly detected if the IoU, defined in Equation~\ref{eq:1}, between it and the ground-truth is larger than $0.5$. Based on these definitions, we can calculate precision (Equation~\ref{eq:2}) and recall (Equation~\ref{eq:3}).
    \begin{equation} \label{eq:2}
    Precision = \frac{TP}{TP + FP}
    \end{equation}
    \begin{equation} \label{eq:3}
    Recall = \frac{TP}{TP + FN}
    \end{equation}
As shown in Figure~\ref{fig:7}, we evenly sampled 21 points as ``foodness" score thresholds in the range [0.0,1.0] and plot precision and recall for each threshold. High precision usually corresponds to low recall. In order to find the optimal combination of precision and recall, F-measure (Equation~\ref{eq:F-M}), which is the harmonic mean of precision and recall, is introduced. In Figure~\ref{fig:7}, we plot F-measure for each threshold and mark the highest value in black.
    \begin{equation} \label{eq:F-M}
    F-measure = 2 \times \frac{Precision \times Recall}{Precision + Recall}
    \end{equation}

As shown in the Table~\ref{tab:1}, we report the selected ``foodness" score threshold and all three performance metrics on the testing set of each food dataset. Precision measures the proportion of positive detection that is actually correct, and recall calculates the proportion of actual positives that is detected correctly. Therefore, we expect our system to have both high values in precision and recall. The VFN datasets has higher precision but relatively low recall. It is worth mentioning that about $6.4\%$ of images in UEC-256 dataset and $9.2\%$ of UEC-100 contain more than 2 bounding boxes, while the VFN dataset has more than $26\%$ images with multiple bounding boxes, making it much more challenging. Due to these multiple food regions, False Negative is higher for the VFN dataset which causes lower recall.

    \begin{table}[ht]

        \centering
        \caption{Performance of food localization on different food image datasets}
        \begin{tabular}{|c|c|c|c|}
            \hline
             Dataset & UEC-100 & UEC-256 & VFN\\
            \hline
            Foodness Score Threshold & 0.35 & 0.55 & 0.80\\
            \hline
            Precision & 0.8159 & 0.9388 & 0.7926\\
            Recall & 0.8604 & 0.8764 & 0.6372\\
            F-Measure & 0.8376 & 0.9065 & 0.7064\\
            \hline
        \end{tabular}
        \label{tab:1}
    \end{table}
    
\subsection{Food Classification}\label{classification_exp}
In our system, classification is performed on food regions identified by food localization. Ideally, each food region should contain only one food item. We selected Food-101, UPMC-101, UEC-100, UEC-256, and VFN to evaluate the performance of proposed food classification. Among these datasets, Food-101 and UPMC-101 contain only single food images.  Since the images in UEC-100, UEC-256, and VFN datasets contain multiple food items, we cropped food regions based on ground-truth bounding box information to generate single food images to evaluated the performance of proposed food classification. 
We follow the selection of training and testing data as provided by the Food-101 and UPMC-101 datasets. For other three datasets, we select $70\%$ for training, $10\%$ for validation, and $20\%$ for testing as standard practice.
We report multi-class accuracy, the percentage of correctly classified images, on the testing set of each dataset.

There are many powerful deep learning models developed for food classification. It is worth mentioning that the use of more complicated Neural Networks models, such as Inception~\cite{szegedy2017inception}, ResNet~\cite{resnet} and DenseNet~\cite{huang2017densely}, can achieve higher accuracy. However, our proposed method is agnostic about the underlying CNN model used, so we want to show the improvement of the classification accuracy based on a hierarchical structure of the class labels. Therefore, the term flat classification is referred to tuning a CNN model for food classification, and hierarchical classification means a hierarchical structure generated by visual semantics among class labels is used for food classification.  

We used DenseNet-121 as our backbone CNN model for this evaluation. The evaluation of classification task consists of three experiments. We trained the selected CNN model for flat classification (CNN-FC) at the learning rate of 0.0001, the same CNN model for hierarchical classification (CNN-HC) with the same learning rate at 0.0001, and fine-tuning the same CNN model for hierarchical classification (CNN-HC-FT) at a smaller learning rate of 0.00001. As discussed earlier, it is possible to replace DenseNet-121 with other CNN models to improve the performance of the baseline method, i.e., flat classification. Our CNN model is initialized with pre-trained weights on ImageNet, and uses batch size of 20. 

To generate the hierarchical structure, we first train the DenseNet-121 for flat classification. Once the accuracy and loss converge, we extract the feature maps from each training image and compute the similarity score (OVL) among all categories to generate clusters using Affinity Propagation. As shown in Table~\ref{tab:cluster}, each dataset has a different number of clusters. We designed a 2-level hierarchical structure based on these clusters, i.e., the $1^{st}$-level (bottom-level) is food category and the $2^{nd}-$level (top-level) is food cluster.

\begin{table}[ht]
\centering
\caption{Category and Cluster Numbers in Different Datasets}
  \label{tab:cluster}
  \begin{tabular}{|c|c|c|c|c|c|}
    \hline
    ~ &Food-101&UPMC-101&UEC-100&UEC-256&VFN\\
    \hline
    Category No.& $101$& $101$&$100$&$256$&$82$\\
    Cluster No.& $17$& $18$&$15$&$33$&$14$\\
    \hline
  \end{tabular}
\end{table}

There are two tasks assigned to our multi-task model, one is to classify the food cluster, and the other is to classify the food category. As shown in the Table \ref{tab:single}, with fixed learning rate and batch size, hierarchical classification achieves better performance for top-1 accuracy. After fine-tuning the multi-task model with a smaller learning rate, the accuracy can be further improved.

\begin{table}[ht]
\centering
\caption{Single Food Image Classification Top-1 Accuracy}
  \label{tab:single}
  \begin{tabular}{|c|c|c|c|c|c|}
    \hline
    ~ &Food-101&UPMC-101&UEC-100&UEC-256&VFN\\
    \hline
    CNN-FC & $0.7531$& $0.6483$&$0.7823$&$0.6708$&$0.6520$\\
    CNN-HC & $0.7625$&$0.6601$&$0.7940$&$0.6857$&$0.6784$\\
    CNN-HC-FT & $0.7978$& $0.6926$&$0.8081$&$0.7236$&$0.7181$\\
    \hline
  \end{tabular}
\end{table}

From a nutrient perspective, visually similar foods may contain similar nutrition content, e.g., fried chicken and fried pork. Therefore, our proposed recognition system can minimize the impact of a mistake by clustering visually similar foods together. It is worth mentioning that top-5 accuracy is another performance metric often used in the classification task. However, top-5 accuracy cannot reflect how good or bad a mistake is. As a result, We define a new performance metric called ``Cluster Top-1" accuracy to measure whether our system made a good or bad mistake. The proposed visual-aware hierarchical structure contains many clusters (top level) that consist of several visually similar foods ($2^{nd}$ level). Therefore, if the top-1 decision is a member of the cluster that the correct category belongs to, we consider it a correct ``Cluster Top-1" decision. As shown in Table~\ref{tab:cluster_acc}, the visual-aware hierarchical structure not only improves the top-1 accuracy, but can also improve the cluster top-1 accuracy. In other words, our proposed system is able to make a better mistake than flat classification methods.
 
\begin{table}[ht]
\centering
\caption{Single Food Image Classification Cluster Top-1 Accuracy}
  \label{tab:cluster_acc}
  \begin{tabular}{|c|c|c|c|c|c|}
    \hline
    ~ &Food-101&UPMC-101&UEC-100&UEC-256&VFN\\
    \hline
    CNN-FC & $0.8506$& $0.7426$&$0.8895$&$0.7839$&$0.7986$\\
    CNN-HC & $0.8512$& $0.7607$&$0.8944$&$0.8049$&$0.8217$\\
    CNN-HC-FT & $0.8782$& $0.7873$&$0.9020$&$0.8337$&$0.8481$\\
    \hline
  \end{tabular}
\end{table}

There are many food images in UEC-100, UEC-256, and VFN datasets contain only a single food. As mentioned in Section~\ref{recognition}, food localization can help remove the non-food background pixels to improve the classification performance. To show this benefit, we selected the single food images from UEC-100, UEC-256 and VFN datasets and compose two new datasets: (1) original images which contain single food and background (2) cropped images which contain food region cropped from single food images without background.
Our split for these two datasets is under the constraint that each training/validation/testing image of (2) is cropped from corresponding image in (1). 
To compare the classification performances of these two dataset, we trained vanilla DenseNet-121 (flat classification) under the experiment settings mentioned at the beginning of this section for these datasets. As shown in Table~\ref{tab:Crop}, cropping the food regions in the image indeed improves the classification accuracy, especially for more complex images such as those in the VFN dataset.

\begin{table}[ht]
\centering
\caption{Classification Accuracy of Original Food Images and Food Regions}
  \label{tab:Crop}
  \begin{tabular}{|c|c|c|c|}
    \hline
    ~ &UEC-100&UEC-256&VFN\\
    \hline
    Original Image & $0.7621$& $0.6354$&$0.5542$\\
    Cropped Image & $0.7847$& $0.6560$&$0.6385$\\
    \hline
  \end{tabular}
\end{table}

\subsection{Food Recognition}
In this section, we report the evaluation of our overall system by combining the food localization and food recognition. 
For single food images, localization can help remove irrelevant background pixels. For multi-food images, localization can detect all food regions. Since UEC-100, UEC-256, and VFN datasets have ground-truth bounding box information, we use their test images to evaluate the performance of our food recognition system.

Since it is impossible for predicted bounding boxes from food localization to match exactly to the ground-truth, we followed the most common standards of practice for recognition task using precision and recall, which are described in Section~\ref{sec:localization_exp}. The difference is that we have multiple category labels instead of food/non-food.
For example, the predicted bounding box is treated as True Positive (TP) when it is assigned the correct food label and the IoU between the predicted and ground-truth bounding box is larger than 0.5.
Precision (Equation~\ref{eq:2}) and recall (Equation~\ref{eq:3}), are calculated the same way as illustrated in Section~\ref{sec:localization_exp}.  For comparison, we also calculate the accuracy (Equation~\ref{eq:10}) defined in~\cite{Bolaos2016SimultaneousFL}, which is different from the definition of multi-class accuracy for classification in Section~\ref{classification_exp}.
    \begin{equation} \label{eq:10}
    Accuracy = \frac{TP}{TP + FP + FN}
    \end{equation}
As shown in Table~\ref{tab:256}, we compare our method with \cite{Bolaos2016SimultaneousFL} on UEC-256 dataset.
Reported precision, recall, and accuracy are solely based on label correctness and IoU metric, they do not consider the confidence score of each bounding box assigned by the localization step. Both precision and recall of our method are much higher than those reported in \cite{Bolaos2016SimultaneousFL}. The reasons for the improvements include (1) the food localization step in our system is fully-supervised by bounding-box level annotations, which is stronger than the image level supervision used in ~\cite{Bolaos2016SimultaneousFL} for region proposal. (2) the classification backbone network used in our system, DenseNet-121, is more powerful than GoogleNet~\cite{szegedy2015going} in ~\cite{Bolaos2016SimultaneousFL}. (3) the hierarchical structure based on visual similarity further improves the classification performance as indicated in Section \ref{classification_exp} .

\begin{table}[ht]
\centering
\caption{Comparison With Other Method on UEC-256 dataset}
  \label{tab:256}
  \begin{tabular}{|c|c|c|c|}
    \hline
    ~ &Precision&Recall&Accuracy\\
    \hline
    Other Method~\cite{Bolaos2016SimultaneousFL}& $0.5433$& $0.5086$&$0.3684$\\
    Our Method& $0.6560$& $0.6124$&$0.4635$\\
    \hline
  \end{tabular}
\end{table}

\begin{table}[ht]
\centering
\caption{Precision, Recall, Accuracy, and mAP of Our Method On Three Datasets}
  \label{tab:PRA}
  \begin{tabular}{|c|c|c|c|c|}
    \hline
    ~ &Precision&Recall&Accuracy&mAP\\
    \hline
    UEC-100& $0.6309$& $0.6654$&$0.4790$&$0.6063$\\
    UEC-256& $0.6560$& $0.6124$&$0.4635$&$0.5673$\\
    VFN& $0.5655$& $0.4546$&$0.3369$&$0.4006$\\
    \hline
  \end{tabular}
\end{table}

We also evaluated our system's performance on all three datasets as shown in Table~\ref{tab:PRA}. To provide a more accurate and fair evaluation, we follow the mean Average Precision (mAP) metric used in PASCAL Visual Object Classes~\cite{voc} and report our experimental results in Table~\ref{tab:PRA}. As we illustrated in Section~\ref{sec:localization_exp}, precision and recall vary as we change the confidence score threshold. Average Precision (AP) for each category is the average precision value for recall value over $0$ to $1$ for each food category, and mAP is the mean value of all APs of all categories. 

\section{Discussion}\label{Discussion}
We evaluated the performance of various components of our system in Section \ref{sec:result}. Table~\ref{tab:cluster} shows that the proposed method of building the two-level hierarchical structure is applicable for different datasets. Table~\ref{tab:single} and Table~\ref{tab:cluster_acc} shows that by using the same CNNs model and the same learning rate, our proposed method can improve both top-1 accuracy and cluster top-1 accuracy. After fine-tuning with smaller learning rate, the result can be further improved. In addition, Table~\ref{tab:Crop} illustrates that even for single food image, food localization can remove non-food background pixels and improve classification accuracy. Localization is particularly useful for complex images with background clutters such as those in the VFN dataset as indicated by the larger improvement. It is worth mentioning that for the results in both Table~\ref{tab:single} and Table~\ref{tab:Crop}, the performance on VFN is lower than other public datasets due to several reasons. First, the VFN dataset contains many visually similar food categories, e.g. milk, ice cream, and yogurt. In addition, many categories in this dataset contain around 100 images, which is far less than other public datasets, e.g. Food-101 has 1000 images per category. In our future work, we plan to expand the VNF dataset by increasing the number of images per category.

Since UEC-100, UEC-256, and VFN datasets contain bounding box information, we use these datasets to test the performance of the overall recognition system. Our proposed system significantly outperforms previous method \cite{Bolaos2016SimultaneousFL} as shown in Table~\ref{tab:256}. Table~\ref{tab:PRA} shows that the mAP of VFN is lower than that of the other two datasets. Since the mAP highly depends on the confidence score of the class label assigned to the food region by the classification, the lower classification result of VFN leads to the low mAP value which is another indication that VFN is a more challenging dataset. Without taking into consideration the confidence score, Table~\ref{tab:PRA} also reports our proposed method achieved high precision, recall and accuracy results on UEC-100 and UEC-256, and a reasonable performance results on VFN dataset. The low recall of the VFN is mainly caused by the errors in food localization, which may not detect all the food regions in images. This is a main challenge of the VFN dataset which contains significantly more multiple food images than the UEC-100 and UEC-256 datasets as described in Table~\ref{tab:dataset_compare}.

\section{Conclusion}\label{Conclusion}
In this paper, we developed a recognition system consisting of a food localization step to detect food regions in an image and a hierarchical food classification step that can cluster visually similar food categories to improve classification performance. 
We also introduced a high quality and challenging dataset, VIPER-FoodNet (VFN) dataset, which is based on the most commonly consumed foods in the United States and the images are sourced from the Internet including social media platforms such that they are closely related to daily life food images. 

Our food recognition system is evaluated on several public datasets, including the new VFN dataset and showed improved performance. We also discussed opportunities to further improve the system performance on challenging dataset such as the VFN. As part of our future work, we will expand the VNF dataset by obtaining additional images for the existing categories, as well as adding more categories to the dataset.

 \bibliographystyle{splncs04}
 \bibliography{madima}

\begin{thebibliography}{10}
\providecommand{\url}[1]{\texttt{#1}}
\providecommand{\urlprefix}{URL }
\providecommand{\doi}[1]{https://doi.org/#1}

\bibitem{fndds2018}
{USDA} food and nutrient database for dietary studies 2015-2016. Agricultural
  Research Service, Food Surveys Research Group, 2018

\bibitem{wweia}
What we eat in america, nhanes 2015-2016

\bibitem{Aguilar2017GrabPA}
Aguilar, E., Remeseiro, B., Bola{\~n}os, M., Radeva, P.: Grab, pay, and eat:
  Semantic food detection for smart restaurants. IEEE Transactions on
  Multimedia  \textbf{20},  3266--3275 (2017)

\bibitem{ahmad2016mobile}
Ahmad, Z., Bosch, M., Khanna, N., Kerr, D.A., Boushey, C.J., Zhu, F., Delp,
  E.J.: A mobile food record for integrated dietary assessment. Proceedings of
  the 2nd International Workshop on Multimedia Assisted Dietary Management pp.
  53--62 (October 2016), amsterdam, Netherlands

\bibitem{aizawa_2013}
Aizawa, K., Maruyama, Y., Li, H., Morikawa, C.: Food balance estimation by
  using personal dietary tendencies in a multimedia {Food Log}. IEEE
  Transactions on Multimedia  \textbf{15}(8),  2176 -- 2185 (December 2013)

\bibitem{Willsense}
Alharbi, R., Pfammatter, A., Spring, B., Alshurafa, N.: Willsense: Adherence
  barriers for passive sensing systems that track eating behavior. Proceedings
  of the 2017 CHI Conference Extended Abstracts on Human Factors in Computing
  Systems pp. 2329--2336 (2017). \doi{10.1145/3027063.3053271}

\bibitem{bay2008}
Bay, H., Ess, A., Tuytelaars, T., Gool, L.V.: Speeded-up robust features
  ({SURF}). Journal of Computer Vision and Image Understanding
  \textbf{110}(3),  346-- 359 (June 2008)

\bibitem{beijbom2015menu}
Beijbom, O., Joshi, N., Morris, D., Saponas, S., Khullar, S.: Menu-match:
  Restaurant-specific food logging from images. 2015 IEEE Winter Conference on
  Applications of Computer Vision pp. 844--851 (2015)

\bibitem{Bolaos2016SimultaneousFL}
Bola{\~n}os, M., Radeva, P.: Simultaneous food localization and recognition.
  2016 23rd International Conference on Pattern Recognition (ICPR) pp.
  3140--3145 (2016)

\bibitem{bossard14}
Bossard, L., Guillaumin, M., Gool, L.V.: Food-101 -- mining discriminative
  components with random forests. Proceedings of European Conference on
  Computer Vision  \textbf{8694},  446--461 (September 2014), {Zurich},
  Switzerland

\bibitem{boushey_2017new}
Boushey, C.J., Spoden, M., Zhu, F.M., Delp, E.J., Kerr, D.A.: New mobile
  methods for dietary assessment: review of image-assisted and image-based
  dietary assessment methods. Proceedings of the Nutrition Society
  \textbf{76}(3),  283--294 (August 2017)

\bibitem{buhrmester2016amazon}
Buhrmester, M., Kwang, T., Gosling, S.D.: Amazon's mechanical turk: A new
  source of inexpensive, yet high-quality data?  (2016)

\bibitem{Casperson2015AMP}
Casperson, S.L., Sieling, J., Moon, J., Johnson, L.K., Roemmich, J.N., Whigham,
  L.D.: A mobile phone food record app to digitally capture dietary intake for
  adolescents in a free-living environment: Usability study  (2015)

\bibitem{chang2011}
Chang, C.C., Lin, C.J.: Libsvm: a library for support vector machines. ACM
  Transactions on Intelligent Systems and Technology  \textbf{2}(3), ~27 (2011)

\bibitem{vireo172}
Chen, J., Ngo, C.W.: Deep-based ingredient recognition for cooking recipe
  retrieval. Proceedings of the 24th ACM international conference on Multimedia
  pp. 32--41 (2016)

\bibitem{chen2009}
Chen, M., Dhingra, K., Wu, W., Yang, L., Sukthankar, R., Yang, J.: Pfid:
  Pittsburgh fast-food image dataset. Proceedings of the IEEE International
  Conference on Image Processing pp. 289--292 (November 2009), {Cairo, Egypt}

\bibitem{Chen:2012aa}
Chen, M., Yang, Y., Ho, C., Wang, S., Liu, S., Chang, E., Yeh, C., Ouhyoung,
  M.: Automatic chinese food identification and quantity estimation.
  Proceedings of SIGGRAPH Asia Technical Briefs pp. 29:1--29:4 (2012),
  {Singapore, Singapore}

\bibitem{Ciocca2015}
Ciocca, G., Napoletano, P., Schettini, R.: Food recognition and leftover
  estimation for daily diet monitoring. New Trends in Image Analysis and
  Processing -- ICIAP 2015 Workshops pp. 334--341 (2015)

\bibitem{cioccaJBHI}
Ciocca, G., Napoletano, P., Schettini, R.: Food recognition: a new dataset,
  experiments and results. IEEE Journal of Biomedical and Health Informatics
  \textbf{21}(3),  588--598 (2017). \doi{10.1109/JBHI.2016.2636441}

\bibitem{cortes1995support}
Cortes, C., Vapnik, V.: Support-vector networks. Machine learning
  \textbf{20}(3),  273--297 (1995)

\bibitem{mixed_dish}
Deng, L., Chen, J., Sun, Q., He, X., Tang, S., Ming, Z., Zhang, Y., Chua, T.S.:
  Mixed-dish recognition with contextual relation networks. Proceedings of the
  27th ACM International Conference on Multimedia pp. 112--120 (2019)

\bibitem{deng2017joint}
Deng, X., Zhang, Y., Yang, S., Tan, P., Chang, L., Yuan, Y., Wang, H.: Joint
  hand detection and rotation estimation using cnn. IEEE transactions on image
  processing  \textbf{27}(4),  1888--1900 (2017)

\bibitem{2017estimating}
Ege, T., Yanai, K.: Estimating food calories for multiple-dish food photos.
  2017 4th IAPR Asian Conference on Pattern Recognition (ACPR) pp. 646--651
  (2017)

\bibitem{voc}
Everingham, M., Van~Gool, L., Williams, C.K.I., Winn, J., Zisserman, A.: The
  pascal visual object classes (voc) challenge. International Journal of
  Computer Vision  \textbf{88}(2),  303--338 (Jun 2010)

\bibitem{fang_2018_ssiai}
Fang, S., Liu, C., Khalid, K., Zhu, F., Boushey, C., Delp, E.J.: ctada: The
  design of a crowdsourcing tool for online food image identification and
  segmentation. Proceedings of the IEEE Southwest Symposium on Image Analysis
  and Interpretation  (April 2018), {Las Vegas, NV}, 2018

\bibitem{fang-nu2019}
Fang, S., Shao, Z., Kerr, D.A., Boushey, C.J., Zhu, F.: An end-to-end
  image-based automatic food energy estimation technique based on learned
  energy distribution images: Protocol and methodology. Nutrients
  \textbf{11}(4), ~877 (2019)

\bibitem{farinella2014benchmark}
Farinella, G.M., Allegra, D., Stanco, F.: A benchmark dataset to study the
  representation of food images. Computer Vision-ECCV 2014 Workshops pp.
  584--599 (2014)

\bibitem{Farinella2015}
Farinella, G.M., Allegra, D., Stanco, F., Battiato, S.: On the exploitation of
  one class classification to distinguish food vs non-food images. New Trends
  in Image Analysis and Processing -- ICIAP 2015 Workshops pp. 375--383 (2015)

\bibitem{FeiFei2005aa}
Fei-Fei, L., Perona, P.: A bayesian hierarchical model for learning natural
  scene categories. IEEE Conference on Computer Vision and Pattern Recognition
  \textbf{2},  524--531 (2005)

\bibitem{frey2007clustering}
Frey, B.J., Dueck, D.: Clustering by passing messages between data points.
  science  \textbf{315}(5814),  972--976 (2007)

\bibitem{girshick2015fast}
Girshick, R.: Fast r-cnn. Proceedings of the IEEE International Conference on
  Computer Vision pp. 1440--1448 (December 2015)

\bibitem{Hassannejad2016}
Hassannejad, H., Matrella, G., Ciampolini, P., De~Munari, I., Mordonini, M.,
  Cagnoni, S.: Food image recognition using very deep convolutional networks.
  Proceedings of the 2Nd International Workshop on Multimedia Assisted Dietary
  Management pp. 41--49 (2016). \doi{10.1145/2986035.2986042}

\bibitem{he2020multitask}
He, J., Shao, Z., Wright, J., Kerr, D., Boushey, C., Zhu, F.: Multi-task
  image-based dietary assessment for food recognition and portion size
  estimation. arXiv preprint arXiv:2004.13188  (2020)

\bibitem{mask_rcnn}
He, K., Gkioxari, G., Dollar, P., Girshick, R.: Mask r-cnn. Proceedings of the
  IEEE International Conference on Computer Vision pp. 2980--2988 (Oct 2017),
  {Venice, Italy}

\bibitem{resnet}
He, K., Zhang, X., Ren, S., Sun, J.: Deep residual learning for image
  recognition. Proceedisng of the IEEE Conference on Computer Vision and
  Pattern Recognition pp. 770--778 (June 2016), {Las Vegas, NV}

\bibitem{He_2016_CVPR}
He, K., Zhang, X., Ren, S., Sun, J.: Deep residual learning for image
  recognition. The IEEE Conference on Computer Vision and Pattern Recognition
  (CVPR)  (June 2016)

\bibitem{hand_crafted2}
{Hoashi}, H., {Joutou}, T., {Yanai}, K.: Image recognition of 85 food
  categories by feature fusion. 2010 IEEE International Symposium on Multimedia
  pp. 296--301 (Dec 2010). \doi{10.1109/ISM.2010.51}

\bibitem{huang2017densely}
Huang, G., Liu, Z., Van Der~Maaten, L., Weinberger, K.Q.: Densely connected
  convolutional networks. Proceedings of the IEEE conference on computer vision
  and pattern recognition pp. 4700--4708 (2017)

\bibitem{Jia20123DLO}
Jia, W., Yue, Y., Fernstrom, J.D., Zhang, Z., Yang, Y., Sun, M.: 3d
  localization of circular feature in 2d image and application to food volume
  estimation. 2012 Annual International Conference of the IEEE Engineering in
  Medicine and Biology Society pp. 4545--4548 (2012)

\bibitem{jiang2017face}
Jiang, H., Learned-Miller, E.: Face detection with the faster r-cnn. IEEE
  International Conference on Automatic Face \& Gesture Recognition pp.
  650--657 (2017)

\bibitem{joutou2009}
Joutou, T., Yanai, K.: A food image recognition system with multiple kernel
  learning. Proceedings of the IEEE International Conference on Image
  Processing pp. 285--288 (October 2009), {Cairo}, Egypt

\bibitem{highly}
Kagaya, H., Aizawa, K.: Highly accurate food/non-food image classification
  based on a deep convolutional neural network. New Trends in Image Analysis
  and Processing -- ICIAP 2015 Workshops pp. 350--357 (2015)

\bibitem{Kagaya}
Kagaya, H., Aizawa, K., Ogawa, M.: Food detection and recognition using
  convolutional neural network. Proceedings of the 22Nd ACM International
  Conference on Multimedia pp. 1085--1088 (2014), {Orlando, Florida, USA}

\bibitem{kawano2014automatic}
Kawano, Y., Yanai, K.: Automatic expansion of a food image dataset leveraging
  existing categories with domain adaptation. Proceedings of European
  Conference on Computer Vision Workshops pp. 3--17 (September 2014), {Zurich,
  Switzerland}

\bibitem{kirkpatrick2014performance}
Kirkpatrick, S.I., Subar, A.F., Douglass, D., Zimmerman, T.P., Thompson, F.E.,
  Kahle, L.L., George, S.M., Dodd, K.W., Potischman, N.: Performance of the
  automated self-administered 24-hour recall relative to a measure of true
  intakes and to an interviewer-administered 24-h recall. The American journal
  of clinical nutrition  \textbf{100}(1),  233--240 (2014)

\bibitem{foodlogA}
Kitamura, K., Yamasaki, T., Aizawa, K.: Foodlog: Capture, analysis and
  retrieval of personal food images via web. Proceedings of the ACM multimedia
  workshop on Multimedia for cooking and eating activities pp. 23--30 (November
  2009), {Beijing, China}

\bibitem{krizhevsky2012imagenet}
Krizhevsky, A., Sutskever, I., Hinton, G.E.: Imagenet classification with deep
  convolutional neural networks. Proceedings of Advances in Neural Information
  Processing Systems pp. 1097--1105 (December 2012)

\bibitem{larsson2002validity}
Larsson, C.L., Westerterp, K.R., Johansson, G.K.: Validity of reported energy
  expenditure and energy and protein intakes in swedish adolescent vegans and
  omnivores. The American journal of clinical nutrition  \textbf{75}(2),
  268--274 (2002)

\bibitem{dlnature}
LeCun, Y., Bengio, Y., Hinton, G.: Deep learning. Nature  \textbf{521},
  436--444 (May 2015)

\bibitem{lin2013network}
Lin, M., Chen, Q., Yan, S.: Network in network. arXiv preprint arXiv:1312.4400
  (2013)

\bibitem{livingstone2004}
Livingstone, M.B.E., Robson, P.J., Wallace, J.M.W.: Issues in dietary intake
  assessment of children and adolescents. British Journal of Nutrition
  \textbf{92},  S213--S222 (October 2004)

\bibitem{Lowe2004}
Lowe, D.: Distinctive image features from scale-invariant keypoints.
  International Journal of Computer Vision  \textbf{2}(60),  91--110 (January
  2004)

\bibitem{recipe1m+}
Marin, J., Biswas, A., Ofli, F., Hynes, N., Salvador, A., Aytar, Y., Weber, I.,
  Torralba, A.: Recipe1m+: A dataset for learning cross-modal embeddings for
  cooking recipes and food images. IEEE transactions on pattern analysis and
  machine intelligence  (2019)

\bibitem{wide}
{Martinel}, N., {Foresti}, G.L., {Micheloni}, C.: Wide-slice residual networks
  for food recognition. 2018 IEEE Winter Conference on Applications of Computer
  Vision (WACV) pp. 567--576 (March 2018). \doi{10.1109/WACV.2018.00068}

\bibitem{structured}
{Martinel}, N., {Piciarelli}, C., {Micheloni}, C., {Foresti}, G.L.: A
  structured committee for food recognition pp. 484--492 (Dec 2015).
  \doi{10.1109/ICCVW.2015.70}

\bibitem{Matsuda:2012ab}
Matsuda, Y., Hoashi, H., Yanai, K.: Recognition of multiple-food images by
  detecting candidate regions. Proceedings of IEEE International Conference on
  Multimedia and Expo pp. 25--30 (July 2012), {Melbourne, Australia}

\bibitem{Miyano2012FoodRD}
Miyano, R., Uematsu, Y., Saito, H.: Food region detection using bag-of-features
  representation and color feature. VISAPP  (2012)

\bibitem{murphy_2015}
Myers, A., Johnston, N., Rathod, V., Korattikara, A., Gorban, A., Silberman,
  N., Guadarrama, S., Papandreou, G., Huang, J., Murphy, K.: {Im2Calories:}
  towards an automated mobile vision food diary. Proceedings of the IEEE
  International Conference on Computer Vision  (December 2015), {Santiago,
  Chile}

\bibitem{poslusna2009misreporting}
Poslusna, K., Ruprich, J., de~Vries, J.H., Jakubikova, M., van't Veer, P.:
  Misreporting of energy and micronutrient intake estimated by food records and
  24 hour recalls, control and adjustment methods in practice. British Journal
  of Nutrition  \textbf{101}(S2),  S73--S85 (2009)

\bibitem{pouladzadeh2015foodd}
Pouladzadeh, P., Yassine, A., Shirmohammadi, S.: Foodd: Food detection dataset
  for calorie measurement using food images. New Trends in Image Analysis and
  Processing -- ICIAP 2015 Workshops  \textbf{9281},  441--448 (2015)

\bibitem{360}
{Qiu}, J., {Lo}, F.P.., {Lo}, B.: Assessing individual dietary intake in food
  sharing scenarios with a 360 camera and deep learning. 2019 IEEE 16th
  International Conference on Wearable and Implantable Body Sensor Networks
  (BSN) pp.~1--4 (May 2019). \doi{10.1109/BSN.2019.8771095}

\bibitem{Ragusa2016}
Ragusa, F., Tomaselli, V., Furnari, A., Battiato, S., Farinella, G.M.: Food vs
  non-food classification. Proceedings of the 2Nd International Workshop on
  Multimedia Assisted Dietary Management pp. 77--81 (2016).
  \doi{10.1145/2986035.2986041}

\bibitem{redmon2016you}
Redmon, J., Divvala, S., Girshick, R., Farhadi, A.: You only look once:
  Unified, real-time object detection. Proceedings of the IEEE conference on
  computer vision and pattern recognition pp. 779--788 (2016)

\bibitem{Redmon_2017_CVPR}
Redmon, J., Farhadi, A.: Yolo9000: Better, faster, stronger. The IEEE
  Conference on Computer Vision and Pattern Recognition (CVPR)  (July 2017)

\bibitem{ren2015faster}
Ren, S., He, K., Girshick, R., Sun, J.: Faster r-cnn: Towards real-time object
  detection with region proposal networks. Proceedings of Advances in Neural
  Information Processing Systems pp. 91--99 (December 2015)

\bibitem{rockett2003evaluation}
Rockett, H.R., Berkey, C.S., Colditz, G.A.: Evaluation of dietary assessment
  instruments in adolescents. Current Opinion in Clinical Nutrition \&
  Metabolic Care  \textbf{6}(5),  557--562 (2003)

\bibitem{rollo2015evaluation}
Rollo, M., Ash, S., Lyons-Wall, P., Russell, A.: Evaluation of a mobile phone
  image-based dietary assessment method in adults with type 2 diabetes.
  Nutrients  \textbf{7}(6),  4897--4910 (2015)

\bibitem{rother2004grabcut}
Rother, C., Kolmogorov, V., Blake, A.: Grabcut: Interactive foreground
  extraction using iterated graph cuts. ACM Transactions on Graphics
  \textbf{23}(3),  309--314 (2004)

\bibitem{ilsvrc_15}
Russakovsky, O., Deng, J., Su, H., Krause, J., Satheesh, S., Ma, S., Huang, Z.,
  Karpathy, A., Khosla, A., Bernstein, M., Berg, A., Li, F.: Imagenet large
  scale visual recognition challenge. International Journal of Computer Vision
  \textbf{115}(3),  211 -- 252 (2015)

\bibitem{recipe1m}
Salvador, A., Hynes, N., Aytar, Y., Marin, J., Ofli, F., Weber, I., Torralba,
  A.: Learning cross-modal embeddings for cooking recipes and food images.
  Proceedings of the IEEE conference on computer vision and pattern recognition
  pp. 3020--3028 (2017)

\bibitem{Shao2019}
Shao, Z., Mao, R., Zhu, F.: {Semi-Automatic Crowdsourcing Tool for Online Food
  Image Collection and Annotation}. 2019 IEEE International Conference on Big
  Data pp. 5186--5189 (Dec 2019)

\bibitem{CNNfood}
Shimoda, W., Yanai, K.: Cnn-based food image segmentation without pixel-wise
  annotation. New Trends in Image Analysis and Processing -- ICIAP 2015
  Workshops pp. 449--457 (2015)

\bibitem{vgg}
Simonyan, K., Zisserman, A.: Very deep convolutional networks for large-scale
  image recognition. arXiv preprint arXiv:1409.1556  (2014)

\bibitem{Singla2016}
Singla, A., Yuan, L., Ebrahimi, T.: Food/non-food image classification and food
  categorization using pre-trained googlenet model. Proceedings of the 2Nd
  International Workshop on Multimedia Assisted Dietary Management pp. 3--11
  (2016). \doi{10.1145/2986035.2986039}

\bibitem{sun2014ebutton}
Sun, M., Burke, L.E., Mao, Z.H., Chen, Y., Chen, H.C., Bai, Y., Li, Y., Li, C.,
  Jia, W.: ebutton: a wearable computer for health monitoring and personal
  assistance. Proceedings of the 51st annual design automation conference
  pp.~1--6 (2014)

\bibitem{Swain1991ColorI}
Swain, M.J., Ballard, D.H.: Color indexing. International Journal of Computer
  Vision  \textbf{7},  11--32 (1991)

\bibitem{szegedy2017inception}
Szegedy, C., Ioffe, S., Vanhoucke, V., Alemi, A.A.: Inception-v4,
  inception-resnet and the impact of residual connections on learning.
  Thirty-first AAAI conference on artificial intelligence  (2017)

\bibitem{szegedy2015going}
Szegedy, C., Liu, W., Jia, Y., Sermanet, P., Reed, S., Anguelov, D., Erhan, D.,
  Vanhoucke, V., Rabinovich, A.: Going deeper with convolutions. Proceedings of
  the IEEE conference on computer vision and pattern recognition pp.~1--9
  (2015)

\bibitem{Tanno2016}
Tanno, R., Okamoto, K., Yanai, K.: Deepfoodcam: A dcnn-based real-time mobile
  food recognition system. Proceedings of the 2Nd International Workshop on
  Multimedia Assisted Dietary Management pp. 89--89 (2016).
  \doi{10.1145/2986035.2986044}

\bibitem{uijlings2013selective}
Uijlings, J.R., van~de Sande, K.E., Gevers, T., Smeulders, A.W.: Selective
  search for object recognition. International journal of computer vision
  \textbf{104}(2),  154--171 (2013)

\bibitem{OVL}
Vijaymeena, M., Kavitha, K.: A survey on similarity measures in text mining.
  Machine Learning and Applications: An International Journal  \textbf{3}(2),
  19--28 (2016)

\bibitem{wang2015recipe}
Wang, X., Kumar, D., Thome, N., Cord, M., Precioso, F.: Recipe recognition with
  large multimodal food dataset. 2015 IEEE International Conference on
  Multimedia \& Expo Workshops (ICMEW) pp.~1--6 (2015)

\bibitem{wang2018context}
Wang, Y., He, Y., Boushey, C.J., Zhu, F., Delp, E.J.: Context based image
  analysis with application in dietary assessment and evaluation. Multimedia
  tools and applications  \textbf{77}(15),  19769--19794 (2018)

\bibitem{Wang2017}
Wang, Y., Zhu, F., Boushey, C.J., Delp, E.J.: {Weakly supervised food image
  segmentation using class activation maps}. 2017 IEEE International Conference
  on Image Processing (ICIP) pp. 1277--1281 (sep 2017).
  \doi{10.1109/ICIP.2017.8296487}

\bibitem{wu2016learning}
Wu, H., Merler, M., Uceda-Sosa, R., Smith, J.R.: Learning to make better
  mistakes: Semantics-aware visual food recognition. Proceedings of the 24th
  ACM international conference on Multimedia pp. 172--176 (2016)

\bibitem{upmc}
{Xin Wang}, {Kumar}, D., {Thome}, N., {Cord}, M., {Precioso}, F.: Recipe
  recognition with large multimodal food dataset. 2015 IEEE International
  Conference on Multimedia Expo Workshops (ICMEW) pp.~1--6 (June 2015).
  \doi{10.1109/ICMEW.2015.7169757}

\bibitem{yanai2015food}
Yanai, K., Kawano, Y.: Food image recognition using deep convolutional network
  with pre-training and fine-tuning. Proceedings of the IEEE International
  Conference on Multimedia \& Expo Workshops pp.~1--6 (July 2015)

\bibitem{yang2010}
Yang, S., Chen, M., Pomerleau, D., Sukhankar, R.: Food recognition using
  statistics of pair-wise local features. Proceedings of the International
  Conference on Computer Vision and Pattern Recognition pp. 2249--2256 (June
  2010)

\bibitem{zagoruyko2016wide}
Zagoruyko, S., Komodakis, N.: Wide residual networks. arXiv preprint
  arXiv:1605.07146  (2016)

\bibitem{zhou2015cnnlocalization}
Zhou, B., Khosla, A., A., L., Oliva, A., Torralba, A.: Learning deep features
  for discriminative localization. Proceedings of the IEEE International
  Conference on Computer Vision and Pattern Recognition  (June 2016)

\bibitem{zhu-2015}
Zhu, F., Bosch, M., Khanna, N., Boushey, C., Delp, E.: Multiple hypotheses
  image segmentation and classification with application to dietary assessment.
  IEEE Journal of Biomedical and Health Informatics  \textbf{19}(1),  377--388
  (January 2015)

\bibitem{zhu2010A}
Zhu, F., Bosch, M., Woo, I., Kim, S., Boushey, C., Ebert, D., Delp, E.J.: The
  use of mobile devices in aiding dietary assessment and evaluation. IEEE
  Journal of Selected Topics in Signal Processing  \textbf{4}(4),  756 --766
  (August 2010)

\end{thebibliography}
\end{document}